\documentclass[10pt,twocolumn,letterpaper]{article}

\pdfoutput=1

\usepackage{cvpr}
\usepackage{times}
\usepackage{epsfig}
\usepackage{graphicx}
\usepackage{amsmath}
\usepackage{amssymb}

\usepackage{verbatim}
\usepackage{textgreek}
\usepackage{multirow}
\usepackage[dvipsnames]{xcolor}
\usepackage[linesnumbered,ruled,vlined]{algorithm2e} 
\usepackage[skip=0pt]{caption}

\usepackage[T1]{fontenc}
\usepackage[utf8]{inputenc}
\usepackage{authblk}
\makeatletter
\renewcommand\AB@affilsepx{ \protect\Affilfont} 
\makeatother

\newcommand{\reb}[1]{\textcolor{black}{#1}}
\newcommand\numberthis{\addtocounter{equation}{1}\tag{\theequation}}

\usepackage{arydshln}
\usepackage[square,sort,comma,numbers]{natbib}  
\usepackage[resetlabels]{multibib}
\newcites{New}{References}

\usepackage[pagebackref=true,breaklinks=true,letterpaper=true,colorlinks,bookmarks=false]{hyperref}

\cvprfinalcopy 

\def\cvprPaperID{7752} 

\ifcvprfinal\fi
\begin{document}
	
	\title{RGBD-Dog: Predicting Canine Pose from RGBD Sensors}
	
	\author[1]{Sin{\'e}ad Kearney
	}
	\author[1]{~~Wenbin Li
	}
	\author[1]{~~Martin Parsons
	}
	\author[2]{~~Kwang In Kim
	}
	\author[1]{~~Darren Cosker
	}
	\affil[1]{University of Bath~~~~~~~~~~~~~~~~~~$^2$UNIST}
	\affil[ ]{\protect\\\tt\small \{s.kearney,w.li,m.m.parsons,d.p.cosker\}@bath.ac.uk~~~~kimki@unist.ac.kr}
	

	\twocolumn[{%
		\renewcommand\twocolumn[1][]{#1}%
		\maketitle
		\vspace{-3\baselineskip}
		\begin{center}
			\centering
			\includegraphics[width=0.99\linewidth]{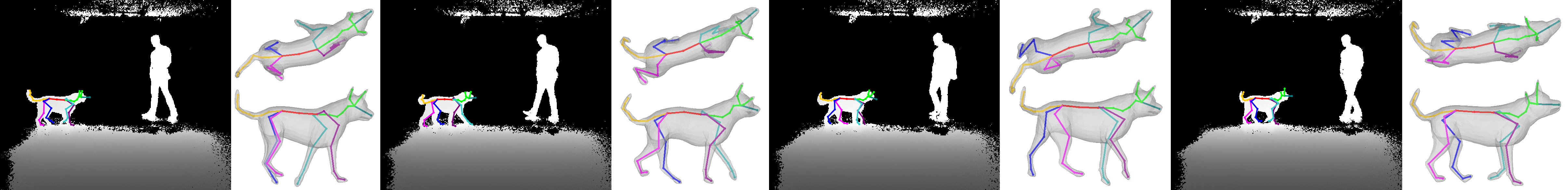}
			\vspace{0.25cm}
			\captionof{figure}{We present a system to predict the skeleton pose of a dog from RGBD images. If the size and shape of the dog is unknown, an estimation is provided. Displayed here are frames [4,7,13,18] of a Kinect sequence, showing 2D projection, 3D skeleton and skinned mesh as produced by the pipeline. All figures in this paper are most informative when viewed in colour.}
		\end{center}%
	}]

	\begin{abstract}
		\vspace{-1\baselineskip}
		\noindent The automatic extraction of animal \reb{3D} pose from images without markers is of interest in a range of scientific fields.
		Most work to date predicts animal pose from RGB images, based on 2D labelling of joint positions.
		However, due to the difficult nature of obtaining training data, no ground truth dataset of 3D animal motion is available to quantitatively evaluate these approaches. 
		In addition, a lack of 3D animal pose data also makes it difficult to train 3D pose-prediction methods in a similar manner to the popular field of body-pose prediction.
		In our work, we focus on the problem of 3D canine pose estimation from RGBD images, recording a diverse range of dog breeds with several Microsoft Kinect v2s, simultaneously obtaining the 3D ground truth skeleton via a motion capture system.
		We generate a dataset of synthetic RGBD images from this data.
		A stacked hourglass network is trained to predict 3D joint locations, which is then constrained using prior models of shape and pose.
		We evaluate our model on both synthetic and real RGBD images and compare our results to previously published work fitting canine models to images.
		Finally, despite our training set consisting only of dog data, visual inspection implies that our network can produce good predictions for images of other quadrupeds -- e.g. horses or cats -- when their pose is similar to that contained in our training set.

	\end{abstract}
	\vspace{-1\baselineskip}

	\section{Introduction}
	\thispagestyle{empty}
	
	\noindent While pose estimation has traditionally focused on human subjects, there has been an increased interest on animal subjects in recent years (\cite{cao2019crossdomain}, \cite{biggs2018creatures}, \cite{zuffi2019threed}, \cite{zuffi2018lions}). It is possible to put markers on certain trained animals such as dogs to employ marker-based motion capture techniques. Nevertheless, there are far more practical difficulties associated with this when compared with human subjects. Some animals may find markers distressing and it is impossible to place them on wild animals. Neural networks currently achieve the best results for human pose estimation, and generally require training on widely available large-scale data sets that provide 2D and/or 3D annotations (\cite{sigal2010humaneva}, \cite{datasetCmu}, \cite{ionescu2013human3}, \cite{Johnson10}). However, there are currently no datasets of 3D animal data available at the same scale concerning the number of samples, variety and annotations, making comparable studies or approaches to pose prediction difficult to achieve. 
	
	In this paper, we propose a markerless approach for 3D skeletal pose-estimation of canines from RGBD images. 
	To achieve this, we present a canine dataset which includes skinned 3D meshes, as well as synchronised RGBD video and 3D skeletal data acquired from a motion capture system which acts as ground truth. 
	Dogs are chosen as our capture subject for several reasons: they are familiar with human contact and so generally accept wearing motion capture suits; they can be brought into the motion capture studio with ease; they respond to given directions producing comparable motions across the numerous subjects; their diverse body shape and size produces data with interesting variations in shape.
	We propose that our resulting dog skeleton structure is more anatomically correct when compared with that of the SMAL model and a larger number of bones in the skeleton allows more expression.
	
	It is challenging to control the capture environment with (uncontrolled) animals - covering wide enough variability in a limited capture session proved to be challenging.
	Hence our method utilises the dog skeletons and meshes produced by the motion capture system to generate a large synthetic dataset.
	This dataset is used to train a predictive network and generative model using 3D joint data and the corresponding projected 2D annotations.
	Using RGB images alone may not be sufficient for pose prediction, as many animals have evolved to blend into their environment and similarly coloured limbs can result in ambiguities. 
	On the other hand, depth images do not rely on texture information and give us the additional advantage of providing surface information for predicting joints. 
	We choose to use the Microsoft Kinect v2 as our RGBD depth sensor, due to its wide availability and the established area of research associated with the device.
	Images were rendered from our synthetically generated 3D dog meshes using the Kinect sensor model of Li et al. \cite{InteriorNet18} to provide images with realistic Kinect noise as training data to the network.
	
	Details of the dataset generation process are provided in Section \ref{sec:synthGen}. Despite training the network with purely synthetic images, we achieve high accuracy when tested on real depth images, as discussed in Section \ref{sec:testRealKinect}. In addition to this, Section \ref{sec:otherAnimals}, we found that training the network only with dogs still allowed it to produce plausible results on similarly rendered quadrupeds such as horses and lions.
	
	
	The joint locations predicted by deep networks may contain errors. In particular, they do not guarantee that the estimated bone lengths remain constant throughout a sequence of images of the same animal and may also generate physically impossible poses. To address these limitations, we adopt a prior on the joint pose configurations -- a Hierarchical Gaussian Process Latent Variable Model (H-GPLVM)~\cite{lawrence2004gaussian}. This allows the representation of high-dimensional non-linear data in lower dimensions, while simultaneously exploiting the skeleton structure in our data. In summary, our main contributions are:
	\begin{itemize}
		\item Prediction of 3D shape \reb{as PCA model parameters}, 3D joint locations and estimation of a kinematic skeleton of canines using RGBD input data.
		\item Combination of a stacked hour glass CNN architecture for initial joint estimation and a H-GPLVM to resolve pose ambiguities, refine fitting and convert joint positions to a kinematic skeleton.
		\item A novel dataset of RGB and RGBD canine data with skeletal ground truth estimated from a synchronised 3D motion capture system and a shape model containing information of both real and synthetic dogs. This dataset and model are available at \footnote{ \url{https://github.com/CAMERA-Bath/RGBD-Dog}.}.
	\end{itemize}

	\section{Related work}
	
	
	\noindent \textbf{2D Animal Pose Estimation}. Animal and insect 2D pose and position data is useful in a range of behavioural studies. Most solutions to date use shallow trained neural network architectures whereby a few image examples of the animal or insect of interest are used to train a keyframe-based feature tracker, e.g. LEAP Estimates Animal Pose \cite{pereira2019fast}, DeepLabCut (\cite{Mathisetal2018}, \cite{nath2019using}) and DeepPoseKit (\cite{graving2019fast}). Cao et al. \cite{cao2019crossdomain} address the issue of the wide variation in interspecies appearance by presenting a method for cross-domain adaption when predicting the pose of unseen species.
	By creating a training dataset by combining a large dataset of human pose (MPII Human Pose \cite{andriluka14cvpr}), the bounding box annotations for animals in Microsoft COCO \cite{lin2014microsoft}, and the authors' animal pose dataset, the method achieves good pose estimation for unseen animals.

	\noindent \textbf{3D Animal Pose Estimation}. Zuffi et al. \cite{zuffi20173d} introduce the Skinned Multi-Animal Linear model (SMAL),
	which separates animal appearance into PCA shape and pose-dependent shape parameters \reb{(e.g. bulging muscles)}, created from a dataset of scanned toy animals.
	A regression matrix calculates joint locations for a given mesh.
	SMAL with Refinement (SMALR) \cite{zuffi2018lions} extends the SMAL model to extract fur texture and achieves a more accurate shape of the animal. 
	In both methods, silhouettes are manually created when necessary, and manually selected keypoints guide the fitting of the model. 
	In SMAL with learned Shape and Texture (SMALST) \cite{zuffi2019threed} a neural network automatically regresses the shape parameters, along with the pose and texture of a particular breed of zebra from RGB images, removing the requirement of silhouettes and keypoints.

	Biggs et al. \cite{biggs2018creatures} fit the SMAL model to sequences of silhouettes that have been automatically extracted from the video using Deeplab \cite{deeplabv3plus2018}.
	A CNN is trained to predict 2D joint locations, with the training set generated using the SMAL model.
	Quadractic programming and genetic algorithms choose the best 2D joint positions.
	SMAL is then fit to the joints and silhouettes.

	In training our neural network, we also generate synthetic RGBD data from a large basis of motion capture data recorded from the real motion of dogs as opposed to the SMAL model and its variants where the pose is based from toy animals and a human-created walk cycle.
	
	\begin{figure*}[h!]
		\begin{center}
			\includegraphics[width=0.9\linewidth]{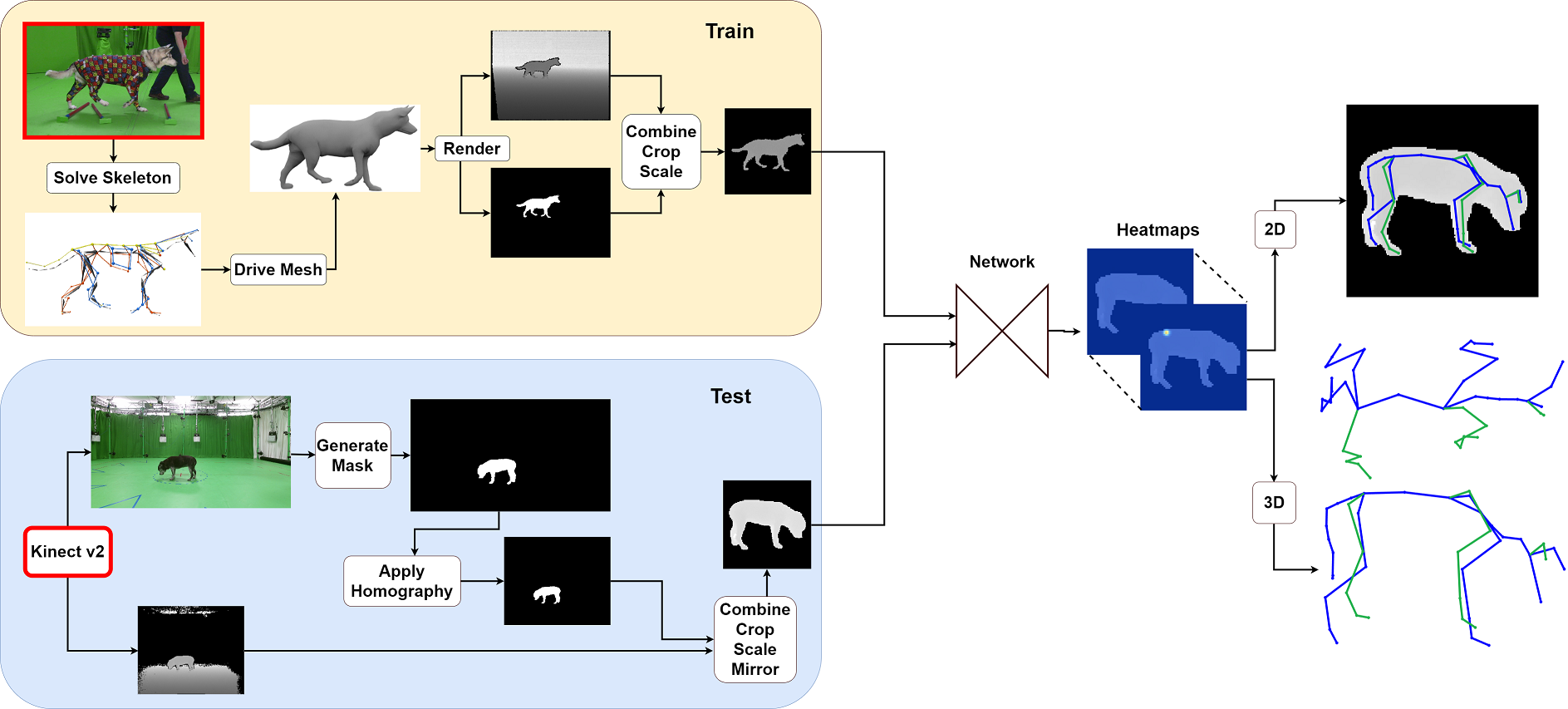}
		\end{center}
		\caption{Overview of the network section of our pipeline. In the training stage, a synthetic dataset is generated from dog motion data. A pair of images is rendered for each frame: depth images are rendered using the Kinect model of InteriorNet \cite{lin2014microsoft} and silhouette masks rendered using OpenGL. In the testing stage, the RGB Kinect image is used to generate a mask of the dog, which is then applied to the depth Kinect image and fed into the network. The network produces a set of 2D heatmaps from which the 2D and 3D joint locations are extracted.}
		\label{fig:overview_network}
	\end{figure*}
	
	\begin{figure*}[h!]
		\begin{center}
			\includegraphics[width=0.9\linewidth]{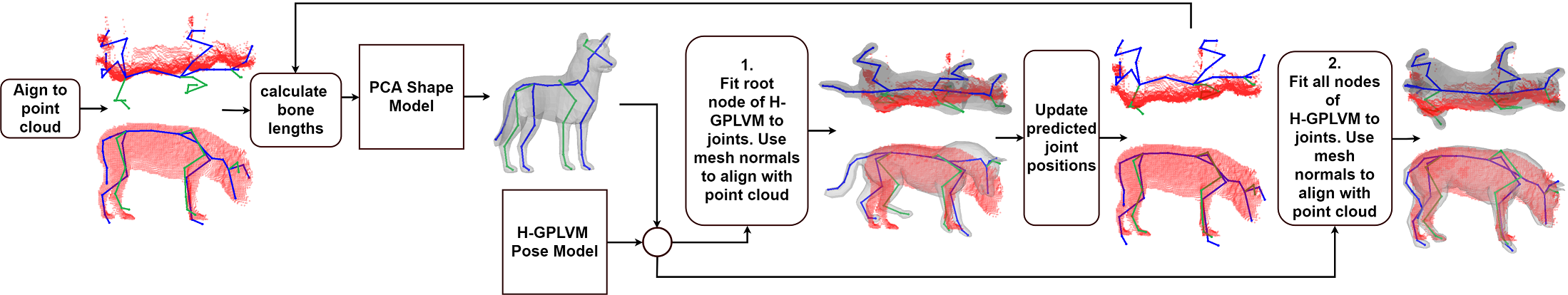}
		\end{center}
		\caption{Overview of the refinement section of our pipeline, showing the steps taken when the dog's neutral body shape is unknown. The point cloud from the depth image initialises the scale of the skeleton and a PCA model predicts body shape from the bone lengths. The H-GPLVM is used to estimate a rough pose of the dog mesh, with the mesh normals then used to refine the mesh/point cloud alignment. The dog scale is refined, the PCA model produces the final shape prediction, and the H-GPVLM fully fits the skinned dog mesh to the point cloud. For known shapes, the PCA prediction steps are not required.}
		\label{fig:overview_refine}
	\end{figure*}
	
	\noindent \textbf{Pose Estimation with Synthetic Training Data}. In predicting pose from RGB images, it is generally found that training networks with a combination of real and synthetic images provides a more accurate prediction than training with either real or synthetic alone (\cite{varol2017learning}, \cite{chen2016synthesizing}, \cite{rogez2016mocap}). Previous work with depth images has also shown that synthetic training alone provides accurate results when tested on real images~\cite{lassner2017generative}. Random forests have been used frequently for pose estimation from depth images.	
	These include labelling pixels with human body parts (\cite{shotton2013real}), mouse body parts (\cite{nanjappa2015mouse}) and dense correspondences to the surface mesh of a human model (\cite{taylor2012vitruvian}). Sharp et al. \cite{sharp2015accurate} robustly track a hand in real-time using the Kinect v2. 

	
	

	
	Recently, neural networks have also been used in pose estimation from depth images.
	Huang \& Altamar \cite{huang2016pose} generate a dataset of synthetic depth images of human body pose and use this to predict the pose of the top half of the body. 
	Mueller et al. \cite{mueller2017real} combine two CNNs to locate and predict hand pose.
	A kinematic model is fit to the 3D joints to ensure temporal smoothness in joint rotations and bone lengths are consistent across the footage.

	
	In our work, we use motion capture data from a selection of dogs to generate a dataset of synthetic depth images. 
	This dataset is used to train a stacked hourglass network, which predicts joint locations in 3D space.
	Given the joints predicted by the network, a PCA model can be used to predict the shape of an unknown dog, and a H-GPLVM is used to constrain the joint locations to those which are physically plausible.
	We believe ours is the first method to train a neural network to predict 3D  animal shape and pose from RGBD images, and to compare our pipeline results to 3D ground truth which is difficult to obtain for animals and has therefore as yet been unexplored by researchers. 
	
	\section{Method}

	\noindent Our pipeline consists of two stages; a prediction stage and refinement stage. In the prediction stage, a stacked hourglass network by Newell et al. \cite{newell2016stacked} predicts a set of 2D heatmaps for a given depth image. From these, 3D joint positions are reconstructed.
	To train the network, skeleton motion data was recorded from five dogs performing the same five actions using a Vicon optical motion capture system (Section \ref{sec:dataCollection}). 
	These skeletons pose a mesh of the respective dog which are then rendered as RGBD images by a Kinect noise-model to generate a large synthetic training dataset (Section \ref{sec:synthGen}).
	We provide more detail about the network training data and explain 3D joint reconstruction from heatmaps in Section \ref{sec:cnn}. In the refinement stage, a H-GPLVM \cite{lawrence2007hierarchical} trained on skeleton joint rotations is used to constrain the predicted 3D joint positions (Section \ref{sec:hgplvm}). The resulting skeleton can animate a mesh, \reb{provided by the user or generated from a shape model,} which can then be aligned to the depth image points to further refine the global transformation of the root of the skeleton. 
	We compare our results with the method of Biggs et al. \cite{biggs2018creatures} and evaluate our method with ground truth joint positions in synthetic and real images in Section \ref{sec:results}. Figures \ref{fig:overview_network} and \ref{fig:overview_refine} outline the prediction and refinement stages of our approach respectively.
	
	\subsection{Animal Motion Data Collection} \label{sec:dataCollection}
\begin{figure}[t]
		\begin{center}
			\includegraphics[width=0.99\linewidth]{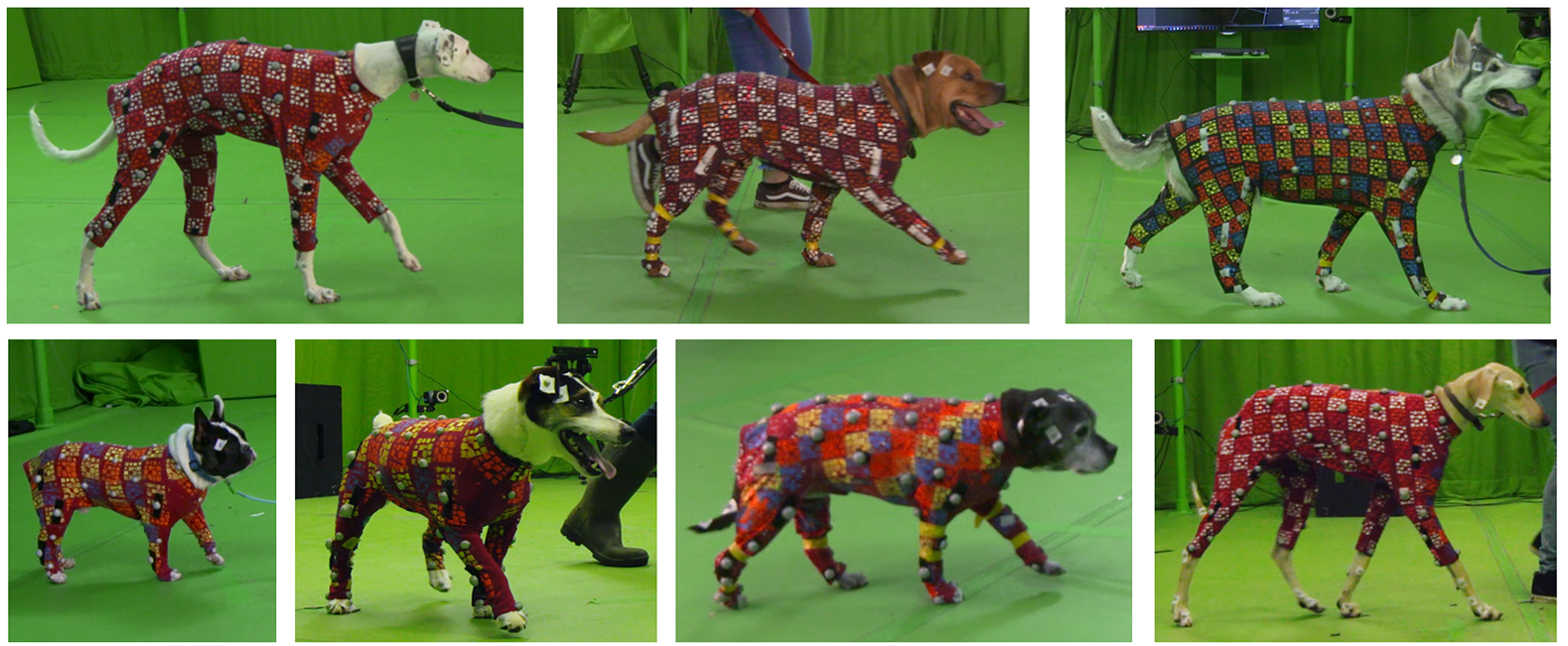}
		\end{center}
		\caption{Dogs included in our dataset, each wearing a motion capture suit. The two dogs on the left were used for test footage only.}
		\label{fig:dogsInSuits}
	\end{figure}

	\noindent As no 3D dog motion data is available for research, we first needed to collect a dataset.
	A local rescue centre provided 16 dogs for recording.
	We focused on five dogs that covered a wide range of shape and size.
	The same five actions were chosen for each dog for the training/validation set, with an additional arbitrary test sequence also chosen for testing.
	In addition to these five dogs, two dogs were used to evaluate the pipeline and were not included in the training set. 
	These dogs are shown in Figure \ref{fig:dogsInSuits}.
	
	A Vicon system with 20 infrared cameras was used to record the markers on the dogs' bespoke capture suits.
	Vicon recorded the markers at 119.88 fps, with the skeleton data exported at 59.94 fps.
	Up to 6 Kinect v2s were also simultaneously recording, with the data extracted using the libfreenect2 library \cite{joshua_blake_2016_45314}. 
	Although the Kinects recorded at 30fps, the use of multiple devices at once reduced overall frame rate  to 6fps in our ground truth set. However, this does not affect the performance of our prediction network. Further details on recording can be found in the supplementary material (Sec. 2.1). 

	\subsection{Synthetic RGBD Data Generation} \label{sec:synthGen}
	
	\noindent Our template dog skeleton is based on anatomical skeletons \cite{dogAnat}.
	Unlike humans, the shoulders of dogs are not constrained by a clavicle and so have translational as well as rotational freedom \cite{dogCollarbone}.
	The ears are modelled with rigid bones and also given translational freedom, allowing the ears to move with respect to the base of the skull.
	In total, there are 43 joints in the skeleton, with 95 degrees of freedom. 
	The neutral mesh of each dog was created by an artist, using a photogrammetric reconstruction as a guide. 
	Linear blend skinning is used to skin the mesh to the corresponding skeleton, with the weights also created by an artist.

\begin{figure}[t]
		\begin{center}
			\includegraphics[width=0.99\linewidth]{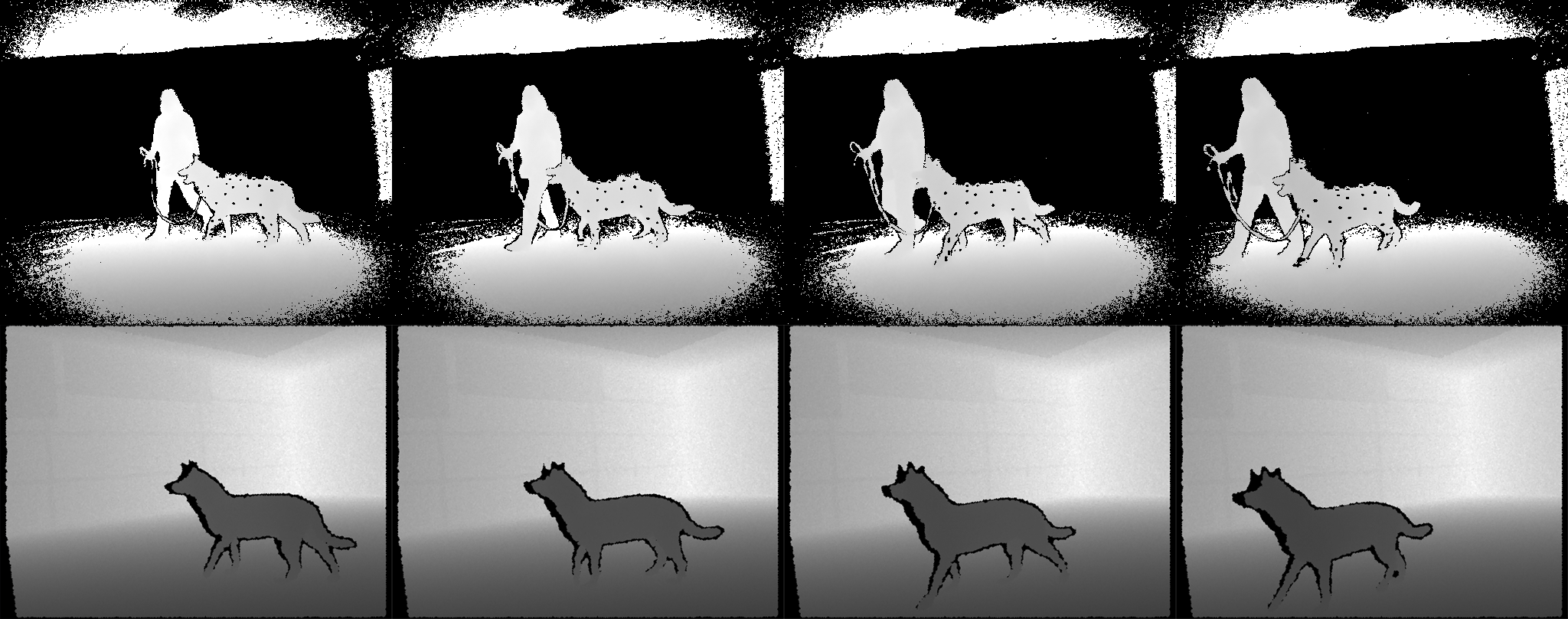}
		\end{center}
		\caption{A comparison of a sequence of real Kinect v2 images (top) with those produced by InteriorNet~\cite{InteriorNet18} (bottom), where all images have been normalised. 
		}
		\label{fig:realVsSynth}
	\end{figure}

	To create realistic Kinect images from our skinned 3D skeletons, we follow a similar process from InteriorNet~\cite{InteriorNet18}. Given a 3D mesh of a dog within virtual environment, we model unique infrared dot patterns projected on to the object, and further achieve dense depth using stereo reconstruction. This process is presumed to retain most of characteristics of Kinect imaging system including depth shadowing and occlusion.
	A comparison of real versus synthetic Kinect images is shown in Figure \ref{fig:realVsSynth}.
	
	Up to 30 synthetic cameras were used to generate the depth images and corresponding binary mask for each dog.	
	Details of the image and joint data normalisation for the generation of ground truth heatmaps are given in the supplementary material.
	The size of the dataset is doubled by using the mirrored version of these images, giving a total number of ~650,000 images in the training set and ~180,000 images in the validation set. An overview of data generation can be seen in the ``Train" section of Figure \ref{fig:overview_network}.
	
	\subsection{Skeleton Pose Prediction Network} \label{sec:cnn}
	\noindent In order to use the stacked-hourglass framework, we represent joints as 2D heatmaps.
	Input to the network are 256x256 greyscale images, where 3D joints $J_{3D256}$ are defined in this coordinate space.
	Given an input image, the network produces a set of 129 heatmaps $H$, each being 64x64 pixels in size.
	Each joint $j$ in the dog skeleton is associated with three heatmaps, the indices of which is known: $h_{j_{XY}}$, $h_{j_{YZ}}$, $h_{j_{XZ}}$, representing the xy-, yz- and xz-coordinates of $j$ respectively.
	\reb{This set provided the most accurate results in our experiments.}
	To produce the heatmaps required to train the network, $J_{3D256}$ are transformed to a 64x64 image coordinates. 
	Let $J_{3D64}$ be these transformed coordinates, where $J_{3D64} = floor(J_{3D256} /4) + 1$.
	We generate 2D gaussians in the heatmaps centred at the xy-, yz- and xz-coordinates of $J_{3D64}$, with a standard deviation of one pixel.
	Inspired by Biggs et al. \cite{biggs2018creatures}, symmetric joints along the sagittal plane of the animal (i.e. the legs and ears) produce multi-model heatmaps. Further technical details on heatmap generation may be found in the suplementary material.
	
	Our neural network is a 2-stacked hourglass network by Newell et al. \cite{newell2016stacked}.
	This particular network was chosen as the successive stages of down-sampling and up-scaling allow the combination of features at various scales.
	By observing the image at global and local scales, the global rotation of the subject can be more easily determined, and the relationship between joints can be utilised to produce more accurate predictions.
	We implement our network using PyTorch, based on code provided by Yang \cite{cnnGit}.
	RMSprop is used as the optimiser, with a learning rate of 0.0025 and batch size 6.
	Our loss function is the MSE between the ground truth and network-generated heatmaps. 
	
	
	\subsubsection{3D Pose Regression from 2D Joint Locations}\label{sec:cnnJointRecon}

	\noindent Given the network-generated heatmaps, we determine the value of  $J_{3D64}$, the location of each joint in the x-, \mbox{y-,} and z-axis in 64x64 image coordinates.
	Each joint $j$ is associated with three heatmaps: $h_{j_{XY}}$, $h_{j_{YZ}}$, $h_{j_{XZ}}$. 
	For joints that produce unimodal heatmaps, the heatmap with the highest peak from the set of $h_{j_{XY}}$, $h_{j_{YZ}}$, $h_{j_{XZ}}$
	determines the value of two of the three coordinates, with the remaining coordinate taken from the map with the second highest peak.

	For joints with multi-modal heatmaps, we repeat this step referring first to the highest peak in the three heatmaps, and then to the second highest peak.
	This process results in two potential joint locations for all joints that form a symmetric pair ($j_{p1}$, $j_{p2}$).
	If the XY position of the predicted coordinate of $j_{p1}$ is within a threshold of the XY position of $j_{p2}$, we assume that the network has erroneously predicted the same position for both joints.
	In this case, the joint with the highest confidence retains this coordinate, and the remaining joint is assigned its next most likely joint.
	
	Once $J_{3D64}$ has been determined, the coordinates are transformed into  $J_{3D256}$. 
	Prior to this step, as in Newell et al. \cite{newell2016stacked}, a quarter pixel offset is applied to the predictions in  $J_{3D64}$.
	We first determine, within a 4-pixel neighbourhood of each predicted joint, the location of the neighbour with the highest value.
	This location dictates the direction of the offset applied.
	The authors note that the addition of this offset increases the joint prediction precision.
	Finally, $J_{3D64}$ is scaled to fit a 256x256 image, resulting in $J_{3D256}$.
	The image scale and translation acquired when transforming the image for network input is inverted and used to transform the xy-coordinates of $J_{3D256}$ into $J_{2Dfull}$, the projections in the full-size image.
	To calculate the depth in camera space for each joint in $J_{3D256}$, the image and joint data normalisation process is inverted and applied.
	$J_{2Dfull}$ is transformed into $J_{3Dcam}$ using the intrinsic parameters of the camera and the depth of each predicted joint.

	\subsection{Pose Prior Model} \label{sec:hgplvm}

	\noindent While some previous pose models represent skeleton rotations using a PCA model, such as the work by Safonova et al. \cite{safonova2004synthesizing}, we found that this type of model produces poses that are not physically possible for the dog.
	In contrast, a Gaussian Process Latent Variable Model (GPLVM) \cite{lawrence2004gaussian} can model non-linear data and allows us to represent our high dimensional skeleton on a low dimensional manifold.
	A Hierarchical GPLVM (H-GPLVM) \cite{lawrence2007hierarchical} exploits the relationship between different parts of the skeleton.
	Ear motion is excluded from the model.
	As ears are made of soft tissue, they are mostly influenced by the velocity of the dog, rather than the pose of other body parts.
	This reduces the skeleton to from 95 to 83 degrees of freedom.
	Bone rotations are represented as unit quaternions, and the translation of the shoulders are defined with respect to their rest position.
	Mirrored poses are also included in the model. The supplementary material contains further technical specifications for our hierarchy (Sec. 2.3).

	We remove frames that contain similar poses to reduce the number of frames included in the training set $S$.
	The similarity of two quaternions is calculated using the dot product, and we sum the results for all bones in the skeleton to give the final similarity value.
	Given a candidate pose, we calculate the similarity between it and all poses in $S$.
	If the minimum value for all calculations is above a threshold, the candidate pose is added to $S$. Setting the similarity threshold to 0.1 reduces the number of frames in a sequence by approximately 50-66\%. The data matrix is constructed from $S$ and normalised.
	Back constraints are used when optimising the model, meaning that similar poses are located in close proximity to each other in the manifold.
	
	\subsubsection{Fitting the H-GPLVM to Predicted Joints}
	
	\noindent A weight is associated with each joint predicted by the network to help guide the fitting of the H-GPLVM. 
	\reb{Information about these weights is given in the supplementary material.}
	To find the initial coordinate in the root node of H-GPLVM, we use k-means clustering to sample 50 potential coordinates. 
	Keeping the root translation fixed, we find the rotation which minimises the Euclidean distance between the \reb{network-predicted joints and the model-generated joints}. 
	The pose and rotation with the smallest error is chosen as the initial values for the next optimisation step.
	
	The H-GPLVM coordinate and root rotation are then refined.
	In this stage, joint projection error is included, as it was found this helped with pose estimation if the network gave a plausible 2D prediction, but noisy 3D prediction.
	The vector generated by the root node of the model provides the initial coordinates of the nodes further along the tree.
	All leaf nodes of the model, root rotation and root translation are then optimised simultaneously.

	During the fitting process, we seek to minimise the distance between joint locations predicted by the network and those predicted by the H-GPLVM: Equation~\ref{eg:hgplvmFit} defines the corresponding loss function: 
	\vspace{-0.25\baselineskip}
	\begin{align}\label{eg:hgplvmFit}
	\mathcal{L}(X,R,&T,t)= \sum_{b=1}^{B} \gamma_{b}\left\|j_{b} - F(X,R,T,t)_{b}\right\|\nonumber\\
	&+\lambda\sum_{b=1}^{B} \gamma_{b}\left\|\Phi(j_{b}) - \Phi(F(X,R,T,t)_{b})\right\|.
	\end{align}
	Here, $B$ is the number of joints in the skeleton, 
	$J = [j_{1}, ..., j_{b}]$ is the set of predicted joint locations from the network, 
	$\Gamma = [\gamma_{1}, ..., \gamma{b}]$ is the set of weights associated with each joint,
	$\Phi$ is the perspective projection function and
	$\lambda$ is the influence of 2D information when fitting the model. 
	Let $X$ be the set of n-dimensional coordinates for the given node(s) of the H-GPLVM and
	$F$ be the function that takes the set $X$, root rotation $R$, root translation $T$, shoulder translations $t$, and produces a set of 3D joints. Figure \ref{fig:overview_refine} shows the result of process.
	
	\section{Evaluation and Results}\label{sec:results}
	\noindent To evaluate our approach, we predict canine shape and pose from RGBD data on a set of five test sequences, one for each dog.
	Each sequence was chosen for the global orientation of the dogs to cover a wide range, both side-views and foreshortened views, with their actions consisting of a general walking/exploring motion.
	In each case we predict shape and pose and compare these predictions to ground truth skeletons as obtained from a motion-capture system (see Section \ref{sec:dataCollection}). More detailed analysis of experiments as well as further technical details of experimental setup -- as well as video results - may be found in the supplementary material.
	
\begin{figure}[t]
		\begin{center}
			\includegraphics[width=0.8\linewidth]{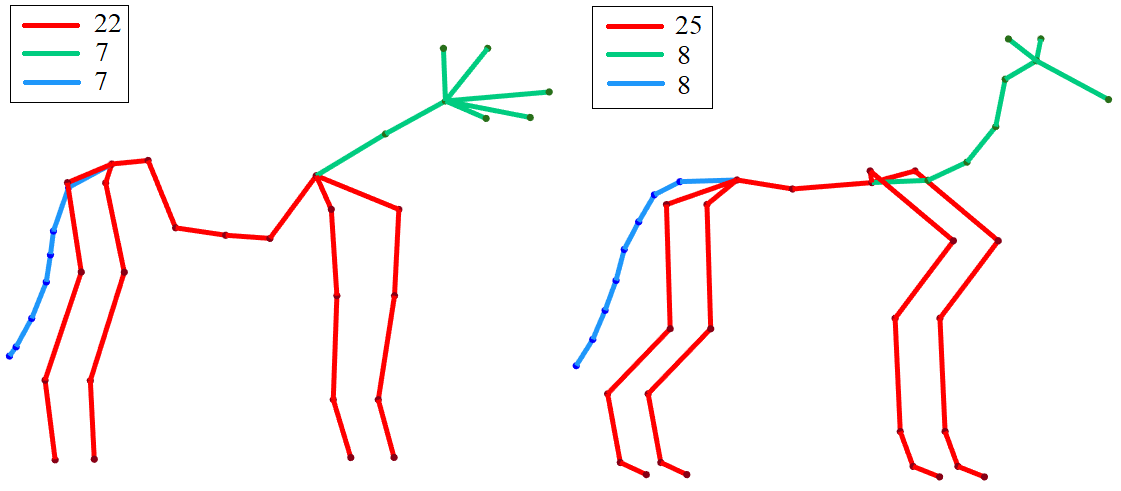}
		\end{center}
		\caption{The number of joints in each skeleton group when evaluating the predicted skeleton against the ground truth skeleton. Left: the SMAL skeleton used by BADJA \cite{biggs2018creatures}, and right: our skeleton. 
		}
		\vspace{-1em}
		\label{fig:skelComp}
	\end{figure}

	\reb{As no previous methods automatically extract dog skeleton from depth images, we compare our results with Biggs et al. \cite{biggs2018creatures}, which we will refer to as the BADJA result.
		We note that the authors' method requires silhouette data only and therefore it is expected that our method produces the more accurate results}.
	Both algorithms are tested on noise-free images.
	We use two metrics to measure the accuracy of our system: Mean Per Joint Position Error (MPJPE) and Probability of Correct Keypoint (PCK). 
	MPJPE measures Euclidean distance and is calculated after the roots of the two skeletons are aligned. 
	A variant PA MPJPE uses Procrustes Analysis to align the predicted skeleton with the ground truth skeleton.
	PCK describes the situation whereby the predicted joint is within a threshold from the true value. 
	The threshold is $\alpha * A$, where $A$ is the area of the image with non-zero pixel values and $\alpha$ = 0.05. 
	The values range from [0,1], where 1 means that all joints are within the threshold.
	PCK can also be used for 3D prediction \cite{mehta2017monocular}, where the threshold is set to half the width of the person's head.
	As we can only determine the length of the head bone, we set the threshold to one and we scale each skeleton such that the head bone has a length of two units.
	To compare the values of MPJPE and PCK 3D, we also use PA PCK 3D, where the joints are aligned as in PA MPJPE, and then calculate PCK 3D. Due to the frequent occlusion of limbs of the dogs, the errors are reported in the following groups: \emph{All} -- all joints in the skeleton; \emph{Head} -- the joints contained in the neck and head; \emph{Body} -- the joints contained in the spine and four legs; -- \emph{Tail}: the joints in the tail. Figure \ref{fig:skelComp} shows the configuration of the two skeletons used and the joints that belong to each group. Our pipeline for each dog contains a separate neural network, H-GPLVM and shape model, such that no data from that particular dog is seen by its corresponding models prior to testing.

	Table \ref{table:badja3d2} contains the PA MPJPE and PA PCK 3D results for the comparison. 
	Comparing these results with the MPJPE and PCK 3D results, for our method, 
	the PA MPJPE decreases the error by an average 0.416 and PA PCK 3D increases by 0.233.
	For BADJA, the MPJPE PA decreases the error by an average 1.557 and PA PCK 3D increases by 0.523, showing the difficulty of determining the root rotation from silhouette alone, as is the case using BADJA.

	\begin{figure}
		\begin{center}
			\includegraphics[width=0.95\linewidth]{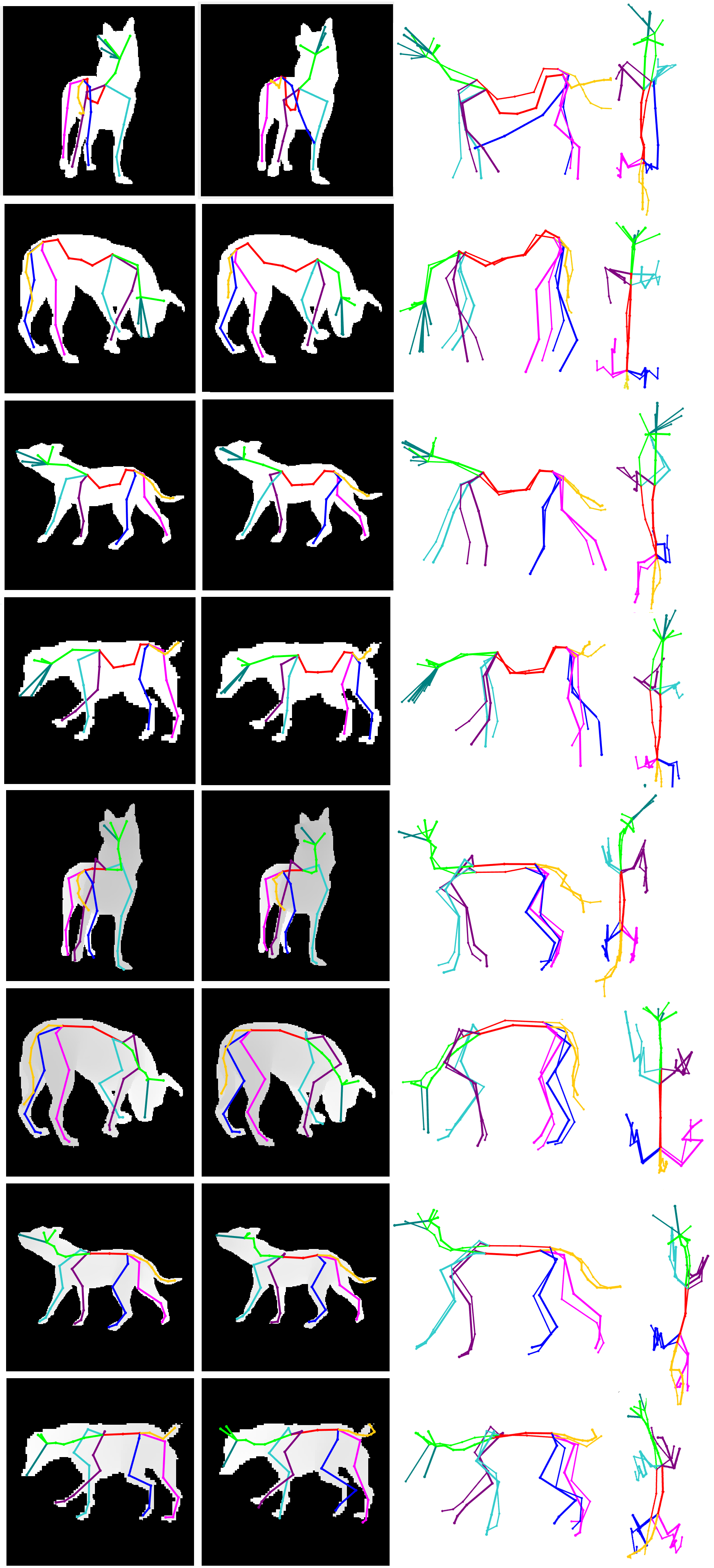}
		\end{center}
		\caption{\reb{An example of results from BADJA \cite{biggs2018creatures} (rows 1-4) and our results (rows 5-8). Column 1 is the ground truth skeleton. Column 2 is the projection of 3D results. Column 3 is a side view of the 3D result as calculated in the PA MJPJE error (where the ground truth shown in a thinner line) and column 4 is a top-down view.}}
		\vspace{-0.5\baselineskip}
		\label{fig:badjaComp_badja}
	\end{figure}

	\begin{table}
		\begin{center}
			\resizebox{\linewidth}{!}{%
				\begin{tabular}{|l|l|l|c|c|c|c|}
					\hline
					Dog & Method & Metric & All & Head  	& Body  & Tail\\
					\hline\hline
					\multirow{4}{*}{Dog1} & \multirow{2}{*}{Ours} & MPJPE & \textbf{0.471} & \textbf{0.382} & \textbf{0.527} & \textbf{0.385} \\
					&	& PCK  & \textbf{0.936} & \textbf{0.984} & \textbf{0.915} & \textbf{0.955} \\
					\cdashline{2-7}
					& \multirow{2}{*}{BADJA\cite{biggs2018creatures}} & MPJPE & 0.976 & 0.993 & 1.002 & 0.879 \\
					&	& PCK & 0.665 & 0.607 & 0.685 & 0.661 \\
					\hline
					\multirow{4}{*}{Dog2} & \multirow{2}{*}{Ours} & MPJPE & \textbf{0.402} & \textbf{0.303} & \textbf{0.410} & \textbf{0.473} \\
					&	& PCK & \textbf{1.000} & \textbf{1.000} & \textbf{1.000} & \textbf{0.998} \\
					\cdashline{2-7}
					& \multirow{2}{*}{BADJA\cite{biggs2018creatures}} & MPJPE & 0.491 &0.392 & 0.524 & 0.486\\
					&	& PCK &  0.956 & \textbf{1.000} & \textbf{1.000} & 0.928\\
					\hline
					\multirow{4}{*}{Dog3} & \multirow{2}{*}{Ours} & MPJPE & \textbf{0.392} & \textbf{0.439} & \textbf{0.390} & \textbf{0.353} \\
					&	& PCK & \textbf{0.985} & \textbf{0.945} & \textbf{0.994} & 0.999 \\
					\cdashline{2-7}
					& \multirow{2}{*}{BADJA\cite{biggs2018creatures}} & MPJPE & 0.610 & 0.843 & 0.617 & 0.356 \\
					&	& PCK & 0.866 & 0.707 & 0.874 & \textbf{1.000} \\
					\hline
					\multirow{4}{*}{Dog4} & \multirow{2}{*}{Ours} & MPJPE & \textbf{0.417} & \textbf{0.395} & \textbf{0.421} & \textbf{0.428} \\
					&	& PCK & \textbf{0.981} & \textbf{0.953} & \textbf{0.985} & \textbf{0.996} \\
					\cdashline{2-7}
					& \multirow{2}{*}{BADJA\cite{biggs2018creatures}} & MPJPE & 0.730 & 0.678 & 0.760 & 0.687 \\
					&	& PCK  & 0.787 & 0.861 & 0.754 & 0.817 \\
					\hline
					\multirow{4}{*}{Dog5} & \multirow{2}{*}{Ours} & MPJPE &\textbf{0.746} & \textbf{0.542} & \textbf{0.748} & 0.944 \\
					&	& PCK & \textbf{0.790} & \textbf{0.925} & \textbf{0.787} & 0.664 \\
					\cdashline{2-7}
					& \multirow{2}{*}{BADJA\cite{biggs2018creatures}} & MPJPE & 0.997 & 0.763 & 1.107 & \textbf{0.885} \\
					&	& PCK & 0.692 & 0.794 & 0.658 & \textbf{0.694}\\
					\hline
			\end{tabular}}
		\end{center}
		\caption{3D error results as calculated using PA MPJPE and PA PCK 3D, comparing our pipeline and that used in BADJA \cite{biggs2018creatures} on each of the 5 dogs. Errors are reported relating to the full body or focussed body parts in Figure \ref{fig:skelComp}.}
		\label{table:badja3d2}
	\end{table}

	\subsection{Applying the Pipeline to Real Kinect Footage}\label{sec:testRealKinect}
	
\begin{figure}[t]
		\begin{center}
			\includegraphics[width=0.9\linewidth]{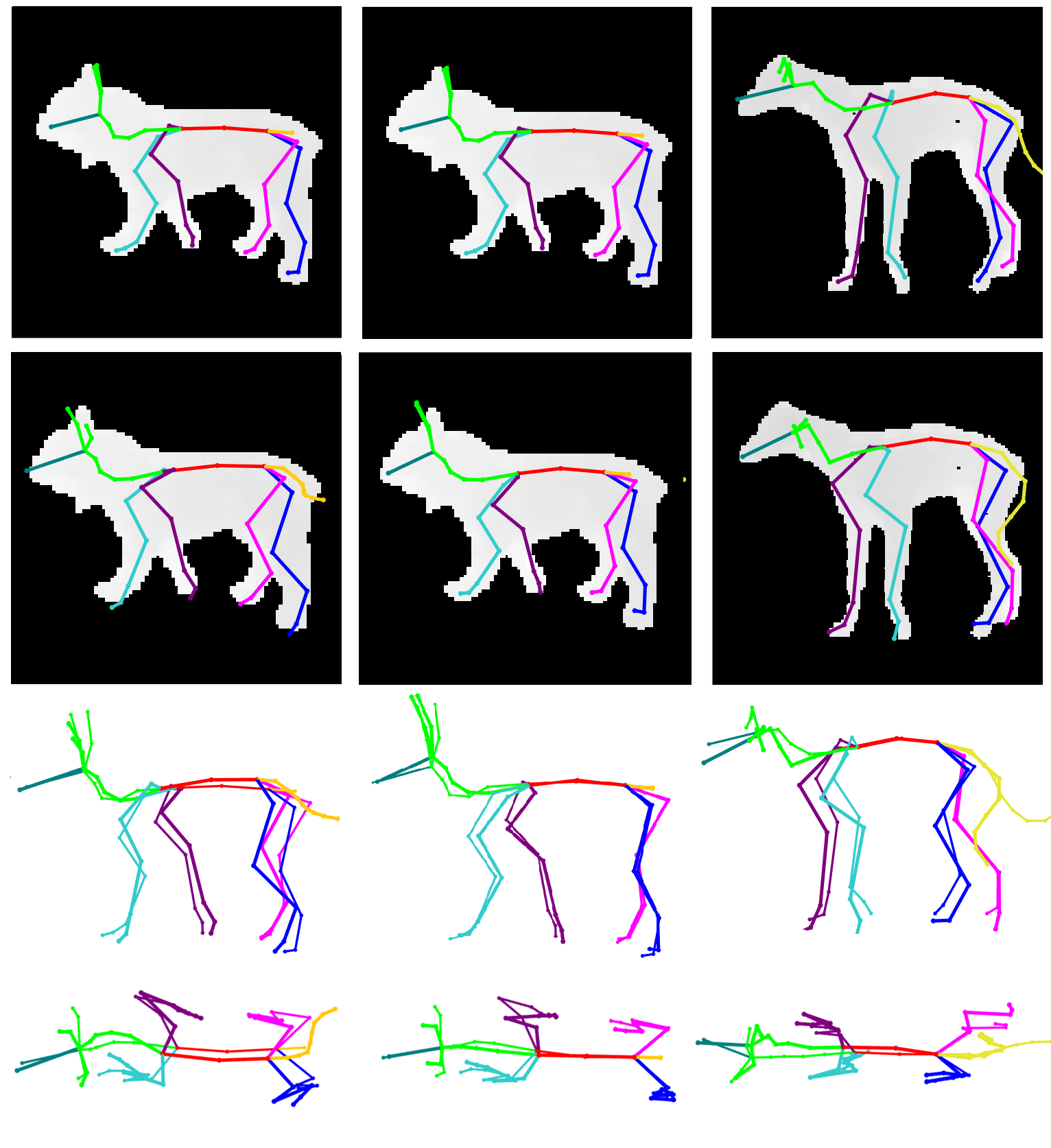}
		\end{center}
		\caption{Example of results on real Kinect images. From the top: ground truth, projection of final 3D result, comparing the 3D result with the thinner ground truth result after calculating PA MPJPE. Left: Dog6, unknown shape. Centre: Dog6, known shape. Right: Dog7, unknown shape.}
		\label{fig:realKinectResult}
		\vspace{-0.5\baselineskip}
	\end{figure}

	\noindent Running the network on real-world data involves the additional step of generating a mask of the dog from the input image. 
	We generate the mask from the RGB image for two reasons: 
	(1) RGB segmentation networks pre-trained to detect animals are readily available,
	(2) the RGB image has a higher resolution than the depth image and contains less noise, particularly when separating the dogs' feet from the ground plane.
	As such, the mask is generated from the RBG image before being transformed using a homography matrix into depth-image coordinates.
	A combination of two pretrained networks are used to generate the mask: Mask R-CNN \cite{he2017mask} and Deeplab \cite{deeplabv3plus2018}.
	More details are included in the supplementary material. We display  3D results in Table \ref{table:realKinect3d}, for cases where the neutral shape of the dog is unknown and known.
	Examples of skeletons are shown in Figure \ref{fig:realKinectResult}.

	\begin{table}
		\begin{center}
			\resizebox{\linewidth}{!}{%
				\begin{tabular}{|l|l|l|c|c|c|c|}
					\hline
					Dog & Method & Metric & All & Head  	& Body  & Tail\\
					\hline\hline
					\multirow{6}{*}{Dog6} & \multirow{2}{*}{CNN} & MPJPE & 0.866  & 0.491 & 0.776 & 1.523 \\
					&	& PCK & 0.745 & 0.956 & 0.780 & 0.425 \\
					\cdashline{2-7}
					& \multirow{2}{*}{H-GPLVM} & MPJPE & 0.667 & 0.466 & 0.627 & 0.993\\
					&	& PCK &  0.873 & 0.969 & 0.938 & 0.575\\
					\cdashline{2-7}
					& H-GPLVM  & MPJPE & \textbf{0.384} & \textbf{0.433} & \textbf{0.437} & \textbf{0.169}\\
					&	(known shape)	& PCK &  \textbf{0.967} & \textbf{0.975} &\textbf{0.954} & \textbf{1.000}\\
					\hline
					\multirow{4}{*}{Dog7} & \multirow{2}{*}{CNN} & MPJPE & 0.563  & \textbf{0.364} & 0.507 & 0.939 \\
					&	& PCK & 0.907 & \textbf{0.993} & 0.943 & 0.707 \\
					\cdashline{2-7}
					& \multirow{2}{*}{H-GPLVM} & MPJPE & \textbf{0.557} & 0.494 & \textbf{0.471} & \textbf{0.888}\\
					&	& PCK &  \textbf{0.922} & 0.947 & \textbf{0.982} & \textbf{0.711}\\
					\hline
			\end{tabular}}
		\end{center}
		\caption{3D Error results of PA MPJPE and PA PCK 3D when using real Kinect images, where each skeleton is scaled such that the head has length of two units. We show the errors for the network prediction (CNN) and the final pipeline result (H-GPLVM). For Dog6, we also show the error where the shape of the dog mesh and skeleton is known.}
		\label{table:realKinect3d}
		\vspace{-0.5\baselineskip}
	\end{table}
	
	\subsection{Shape Estimation of Unknown Dogs}
	\noindent If the skeleton and neutral mesh for the current dog is unknown beforehand -- as is the case in all our results apart from the 'known shape' result in Table 2 -- a shape model is used to predict this information. 
	The model is built from 18 dogs: five dogs are used to train the CNN and were created by an artist, an additional six dogs were also created by the artist, three dogs are scans of detailed toy animals, and four are purchased photogrammetry scans. 
	All dogs are given a common pose and mesh with a common topology. The PCA model is built from the meshes, bone lengths and the joint rotations required to pose the dog from the common pose into its neutral standing position.
	The first four principal components of the model are used to find the dog with bone proportions that best match the recorded dog.
	This produces the estimated neutral mesh and skeleton of the dog.
	
	\subsection{Extending to Other Quadruped Species}\label{sec:otherAnimals}
\begin{figure}[t]
		\begin{center}
			\includegraphics[width=1\linewidth]{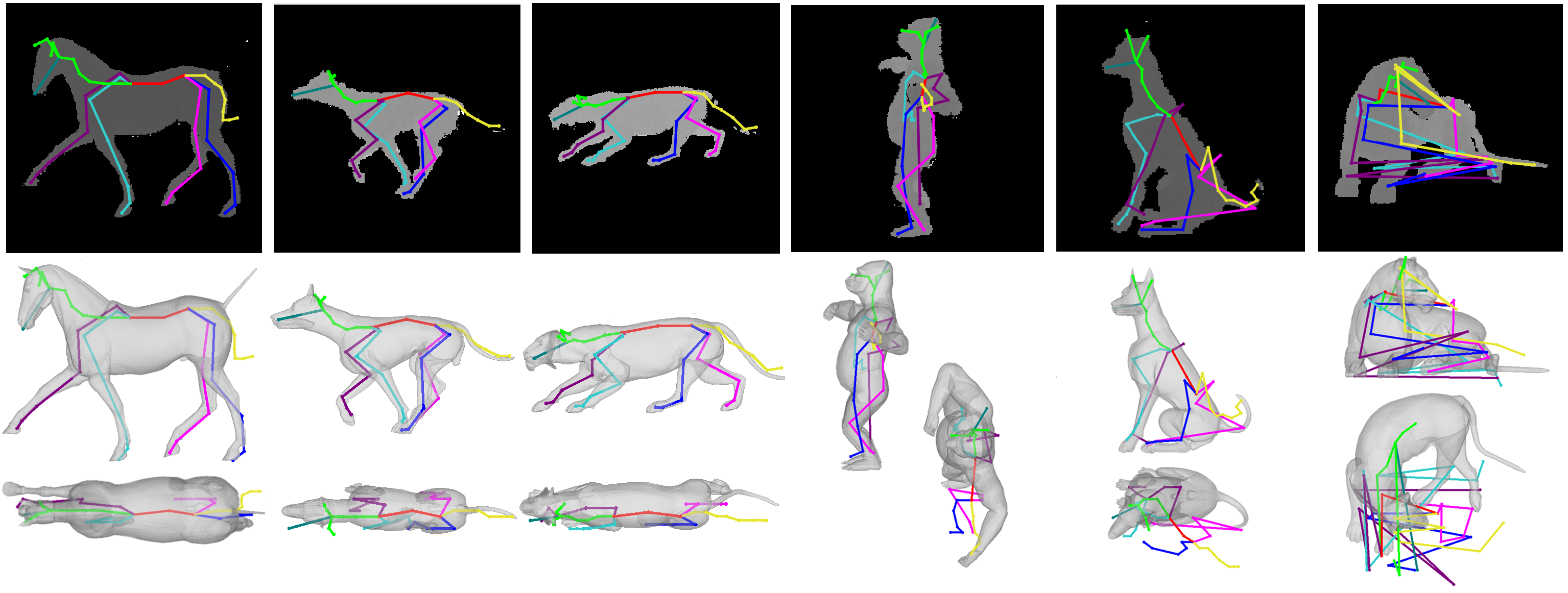}
		\end{center}
		\caption{The network result when given images of a subset of 3D models provided by Bronstein et al. (\cite{bronstein2006efficient}, \cite{bronstein2007calculus}), rendered as in Sec \ref{sec:synthGen}. Although the network is trained with only dog images, the first three columns show the network can generate a good pose for images where the animal is similar to that in the training set. The last three columns show where the network failed to predict a plausible pose.}
		\label{fig:toscaResultGoodBad}
		\vspace{-1em}
	\end{figure}

	\noindent We tested our network on additional 3D models of other species provided by Bronstein et al. (\cite{bronstein2006efficient}, \cite{bronstein2007calculus}).
	Images of the models are rendered as described in Section \ref{sec:synthGen}.
	The training data for the network consists of the same five motions for the five training dogs.
	As no ground truth skeleton information is provided for the 3D models, we evaluate the performance based on visual inspection.
	The example results provided in the first three columns of Figure \ref{fig:toscaResultGoodBad} show that the network performs well when the pose of a given animal is similar to that seen in the training set, even if the subject is not a dog.
	However, when the pose of the animal is very different from the range of poses in the training set, prediction degrades, as seen in the last three columns of Figure \ref{fig:toscaResultGoodBad}. This provides motivation for further work.

	\section{Conclusion and Future Work}
	\noindent We have presented a system which can predict 3D shape and pose of a dog from depth images. 
	We also present to the community a data set of dog motion from multiple modalities - motion capture, RGBD and RGB cameras -- of varying shapes and breeds. 
	Our prediction network was trained using synthetically generated depth images leveraging this data and is demonstrated to work well for 3D skeletal pose prediction given real Kinect input. We evaluated our results against 3D ground truth joint positions demonstrating the effectiveness of our approach. Figure \ref{fig:toscaResultGoodBad} shows the potential in extending the pipeline to other species of animals. We expect that a more pose-diverse training set would produce results more accurate than the failure cases in Figure \ref{fig:toscaResultGoodBad}.
	Apart from the option to estimate bone length over multiple frames, our pipeline does not include temporal constraints, which would lead to more accurate and smoother predict sequences of motion.
	At present, mask generation requires an additional pre-processing step and is based on the RGB channel of the Kinect. Instead, the pose-prediction network could perform a step where the dog is extracted from the depth image itself.
	This may produce more robust masks, as extraction of the dog would no longer rely on texture information.
	As General Adversarial Networks (GANs) are now considered to produce state-of-the-art results, we intend to update our network to directly regress joint rotations and combine this with a GAN to constrain the pose prediction.

	
	\noindent\emph{Acknowledgement.} 
	This work was supported by the Centre for the Analysis of Motion, Entertainment Research and Applications (EP/M023281/1), the EPSRC Centre for Doctoral Training in Digital Entertainment (EP/L016540/1) and the Settlement Research Fund (1.190058.01) of the Ulsan National Institute of Science \& Technology.
	

	{\small
		\bibliographystyle{ieee_fullname}
		\bibliography{egbib}
	}

	\vfill\eject 
	
	\pagebreak
	
	\twocolumn[{%
		\renewcommand\twocolumn[1][]{#1}%
		\vskip .5in
		\begin{center}
			\textbf{\Large RGBD-Dog: Predicting Canine Pose from RGBD Sensors}\\
			\vspace*{4pt}
			\textbf{\Large (Supplementary Material)}\\
			\vspace*{32pt}
			{\large
				Sin{\'e}ad Kearney$^{~1}$ ~Wenbin Li$^{~1}$ ~Martin Parsons$^{~1}$ ~Kwang In Kim$^{~2}$ ~Darren Cosker$^{~1}$\\
			}
			\vskip .5em
			{\large ${}^1$University of Bath~~~~~~~~~~~~~~~~~~$^2$UNIST}
			{\protect\\\tt\small \{s.kearney,w.li,m.m.parsons,d.p.cosker\}@bath.ac.uk~~~~kimki@unist.ac.kr}
		\end{center}
	}]
	\vspace*{4pt}
	
	\setcounter{equation}{0}
	\setcounter{figure}{0}
	\setcounter{table}{0}
	\setcounter{section}{0}
	\setcounter{page}{1}

	\renewcommand{\arraystretch}{1.1} 
	\section{Introduction}

	\noindent In this supplementary material we give additional technical details on our approach.
	We provide details on our dog data set, which will be made available to the research community.
	We conduct additional experiments to test the pipeline for occlusions, exploiting depth information when solving for the shape of the dog, and compare the neural network in the pipeline with two other networks.
	Finally we compare the expression of our dog shape model with that of SMAL \citeNew{zuffi20173dSUPP}. 
	
	\section{Method}
	
	\subsection{ Animal Motion Data Collection}
\begin{figure}
		\begin{center}
			\includegraphics[width=0.8\linewidth]{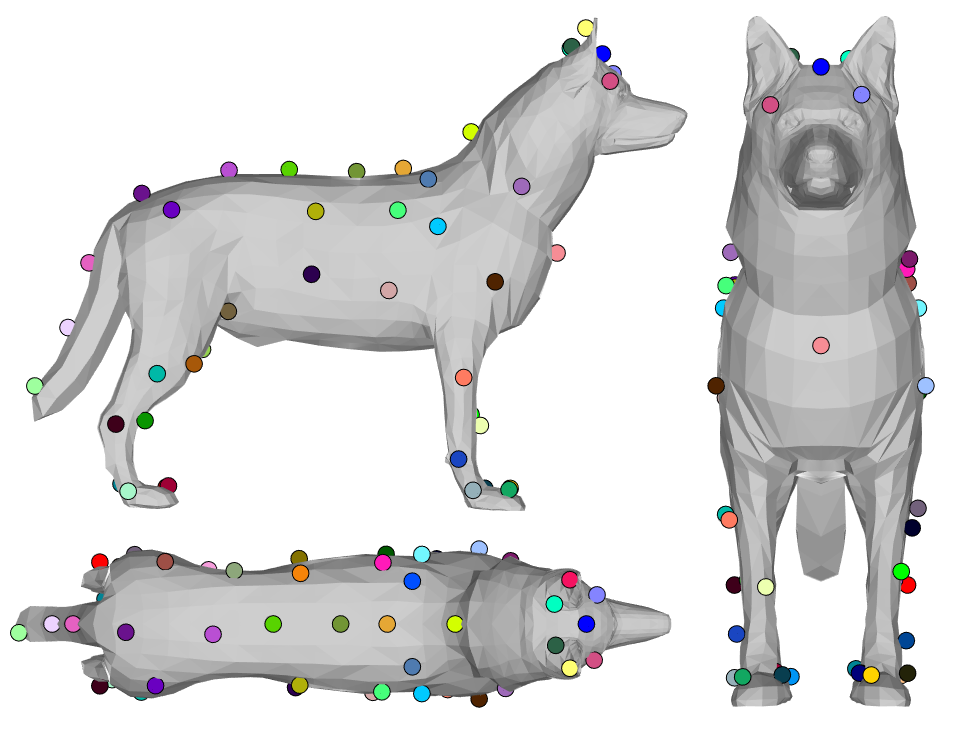}
		\end{center}
		\caption{The locations of the markers as worn by one of the dogs in the capture session\reb{, placed on the artist-created mesh of the dog}. This particular dog had 64 markers in total. Clockwise from top-left: side view, front view, top-down view.}
		\label{fig:markerLoc}
		\vspace{-0.25\baselineskip}
	\end{figure}

\begin{figure}
		\begin{center}
			\includegraphics[width=0.99\linewidth]{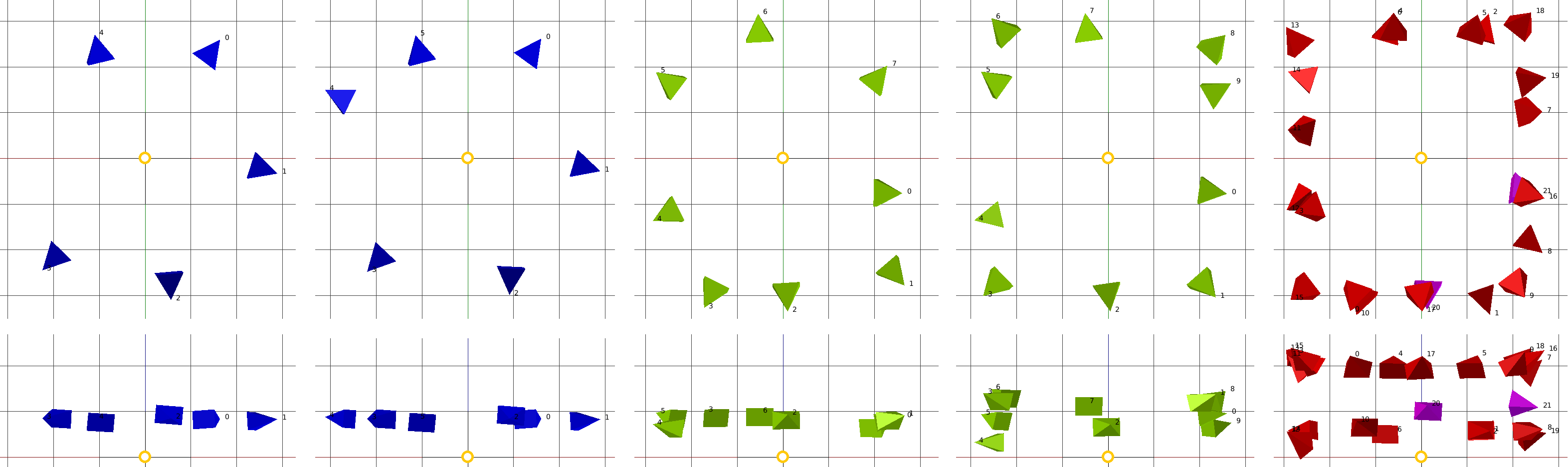}
			
		\end{center}
		\caption{
			The layout of the different camera systems used. Each column is a top-down view (top), and a side-view of the cameras (bottom). Each system is assigned a colour: Kinect cameras are shown in blue, Sony 4K RGB in green and the Vicon cameras in red (the two Vicon witness cameras are shown in magenta). The world origin is denoted with a yellow circle and each grid is 1 metre in width/height/depth. From left: 5-Kinect setup, 6-Kinect setup, 8-Sony setup, 10-Sony setup, the Vicon setup.
		}
		\label{fig:captureSpace}
	\end{figure}

	\noindent Each recorded dog wore a motion capture suit\reb{, on which was painted additional texture information.}
	The number of markers on the suit related to the size of a given dog, and ranged from 63 to 82 markers.
	We show an example of marker locations in Figure \ref{fig:markerLoc}.
	\reb{These locations were based on reference to those on humans and biological study. 
		Vicon Shogun was used to record the dogs with 20 cameras at 119.88fps, the locations of which are shown in Figure \ref{fig:captureSpace}. }
	
\begin{figure*}
		\begin{center}
			\includegraphics[width=0.99\linewidth]{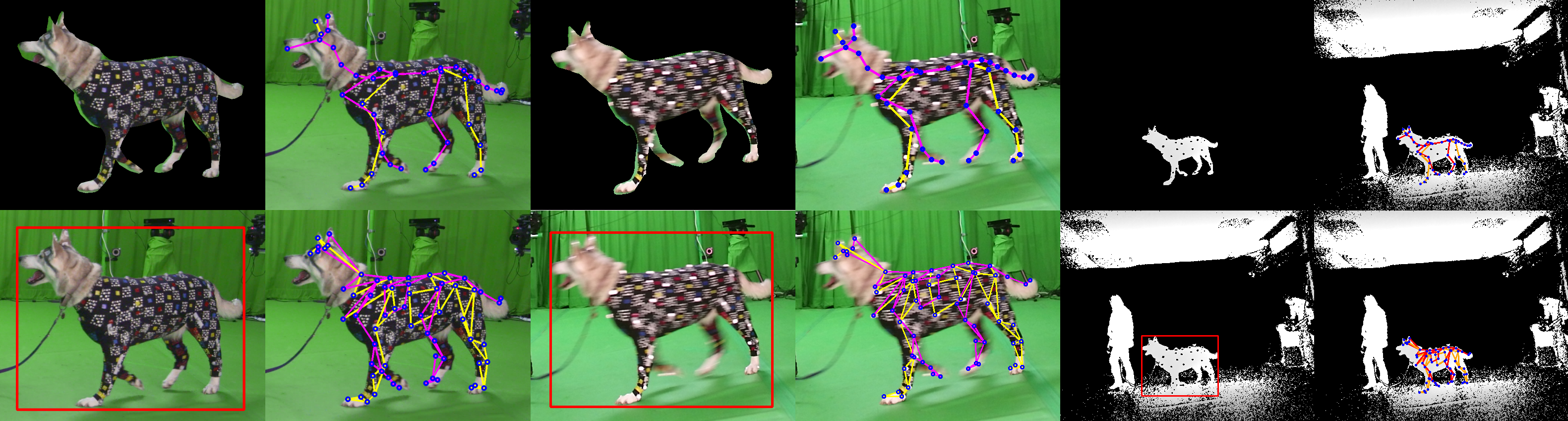}
		\end{center}
		\caption{Image data included in this dataset is, from left, 4K RGB frames, 2K RGB frames from a Microsoft Kinect, and the depth information from a Kinect. Here, the RGB footage is cropped near to the dog bounding box, and the depth image is shown as the full frame. Clockwise from the top left image of each format, we show the image where the silhouette mask has been applied, the projected skeleton of the dog, the projected marker positions of the dogs with connecting lines for ease of identification, and the dog bounding box. For the projection of skeleton and markers in RGB images, yellow denotes limbs on the left side of the body and magenta on the right. For depth images, these colours are orange and red respectively.
		}
		\label{fig:projVals}
	\end{figure*}

	For each dog, this data is available in the form of 3D marker locations, the solved skeleton, the neutral mesh of the dog and corresponding Linear Blend Skinning weights, multi-view HD RGB footage recorded at 59.97 fps, and multi-view RGB and RGB-D images from the Microsoft Kinect recording at approximately 6 fps.
	The HD RGB footage will be available in 4K resolution on request.
	The number of cameras used per dog varied between eight to ten for the HD RGB cameras and five to six for the Kinects.
	Visualisation of this data can be seen in Figure \ref{fig:projVals}.
	The frame count for each sequence of each dog is given in Table \ref{table:frameCount}.
	
	The number of real Kinect RGB and depth images recorded from all cameras for all five motions of the five dogs is 1,950. 
	The number of 4K RGB images recorded from all cameras for all five motions of the five dogs is 73,748.
	In total, 8,346 frames of skeleton motion data were recorded using the Vicon Shogun software.
	
	The number of real Kinect RGB and depth images recorded from all cameras for all five motions of the five dogs is 1,950. 
	The number of 4K RGB images recorded from all cameras for all five motions of the five dogs is 73,748.
	In total, 8,346 frames of skeleton motion data were recorded using the Vicon Shogun software.
	
	In comparison with other available datasets of skeleton-annotated dog images, Biggs et al. \citeNew{biggs2018creaturesSUPP} provide 20 landmarks for 218 frames from video sequences. The size of the images are either 1920x1080 or 1280x720 pixels.
	Cao et al. \citeNew{cao2019crossdomainSUPP} provide 20 landmarks for 1,771 dogs in 1,398 images of various sizes.
	
	\begin{figure}
		\begin{center}
			\includegraphics[width=0.99\linewidth]{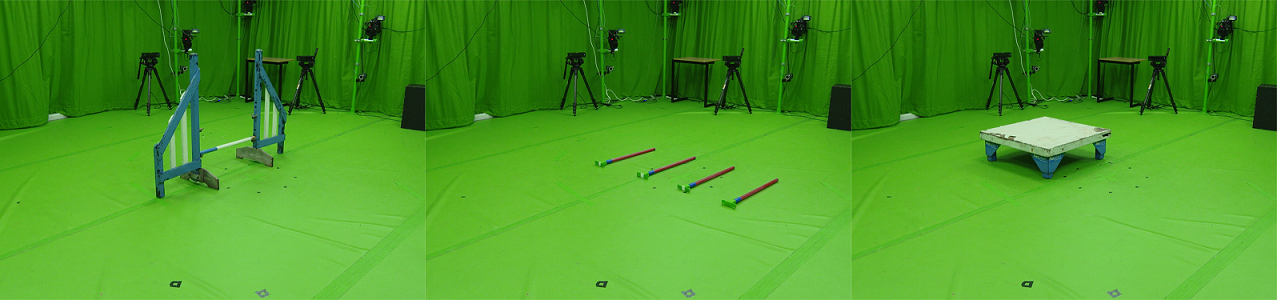}
		\end{center}
		\vspace{-0.25\baselineskip}
		\caption{\reb{The props used during the acquisition of the dataset}}
		\label{fig:props}
	\end{figure}
	
	\begin{table}
		\begin{center}
			\resizebox{\linewidth}{!}{%
				\begin{tabular}{|l|c|c|c|c|c|c|}
					
					\hline
					\multirow{2}{*}{Dogs} & \multicolumn{6}{|c|}{Average \# Frames per Camera (Vicon,Kinect)}\\	
					&    Walk &    Trot &    Jump &   Poles &  Table  & Test \\	
					\hline\hline
					Dog1 & (500,26) & (148,8) & (148,7) & (536,34) & (704,49) & (602,32)
					\\	
					\hline
					Dog2 & (300,39) & (118,7) & (220,19) & (374,50) & (330,24) & (624,52)
					\\	
					\hline
					Dog3 & (152,0) & (138,0) & (232,12) & (232,25) & (582,50)& (602,0)
					\\
					\hline
					Dog4 & (322,0) & (290,0) & (132,0) & (642,0) & (390,0) & (602,0)
					\\	
					\hline
					Dog5 & (596,0) & (188,0) & (324,0) & (376,0) & (372,0) & (602,0)
					\\	
					\hline
					Dog6 & (0,0) & (0,0) & (0,0) & (0,0) & (0,0) & (20,20) \\
					\hline
					Dog7 & (0,0) & (0,0) & (0,0) & (0,0) & (0,0) & (38,38) \\
					\hline
			\end{tabular}}
		\end{center}
		\caption{A table displaying the average number of frames per camera per motion. The first value in each pair refers to the frames from the Vicon system while the second value refers to the Kinect system. For each of the dogs in Dog1-Dog5, the same 5 motions are provided. A separate arbitrary test sequence is also provided if available. For the test dogs, Dog6 and Dog7, only a test sequence is provided.}
		\label{table:frameCount}
	\end{table}
	
	\subsection{Data Augmentation}
	\noindent Our synthetic dataset is generated from the result of applying the processed skeleton motion to the neutral mesh using linear blend skinning.
	The same 20 virtual cameras were used to generate synthetic images for all five dogs, along with cameras using the extrinsic parameters of the 8 to 10 Sony RGB cameras used to record each dog.
	For each motion, two sets of images were generated.
	In the first set, the root of the skeleton contains the rotation and translation of the dog in the scene, and the second set of images are generated where the root has fixed rotation and translation.
	Another version of the two sets was created by mirroring the images, giving our final synthetic dataset approximately 834,650 frames.

	\subsection{\reb{Network Architecture}}
\begin{figure*}
		\begin{center}
			\includegraphics[width=0.99\linewidth]{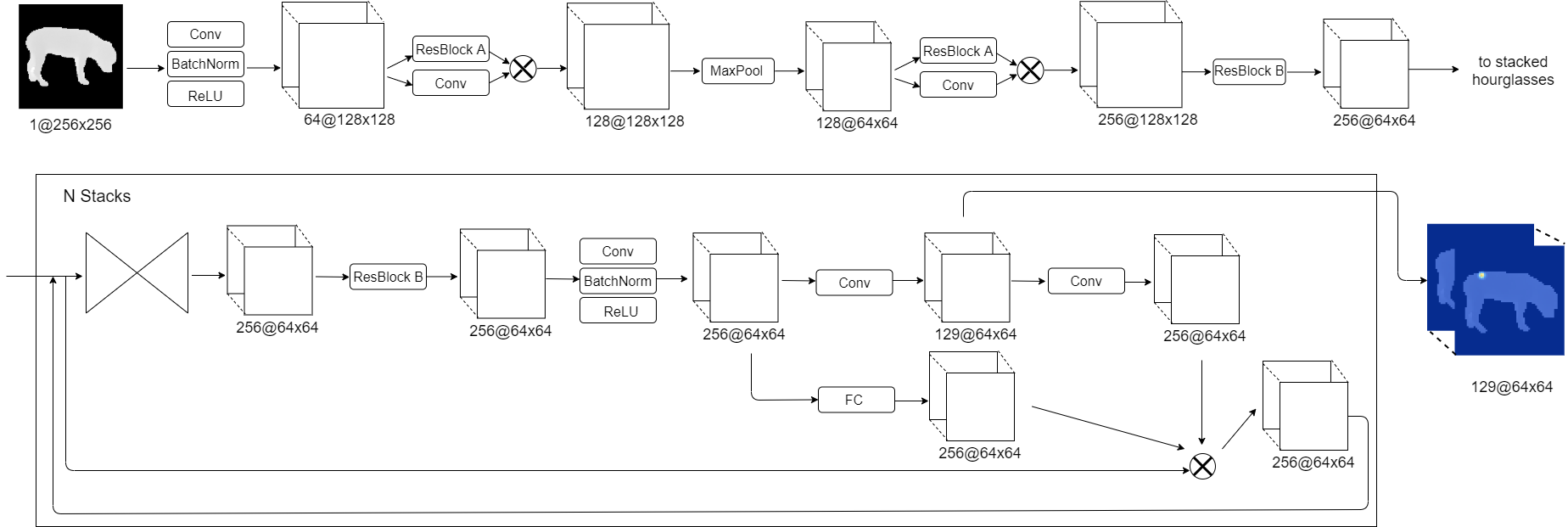}
		\end{center}
		\caption[]{We use the stacked-hourglass network of Newell et al. \citeNew{newell2016stackedSUPP}. In our experiments a stack of two hourglasses is used.``Conv'' stands for convolution and ``FC'' for fully connected. For full details on the network implemented, we direct the user to the paper of Newell et al. \citeNew{newell2016stackedSUPP}.
		}
		\label{fig:network}
	\end{figure*}

	\noindent We use the network of Newell et al. \citeNew{newell2016stackedSUPP} based on the implementation provided by Yang \citeNew{cnnGitSUPP}.
	We provide a diagram of the network in Figure \ref{fig:network} and direct the user to the paper by Newell et al. \citeNew{newell2016stackedSUPP} for full details of the network components.
	
	\subsection{Data Normalisation for the Generation of Training Heatmaps}
	\noindent 3D Joint locations of the skeletons are defined in camera space $J_{3Dcam}$ and 2D joint locations, $J_{2Dfull}$ are their projected values in the synthetic image.
	Only images where all joints in $J_{2Dfull}$ are within the image bounds were included in the dataset.
	The images are shaped to fit the network inputs by following the steps outlined in Algorithm \ref{al:imCrop}, producing images that are 256x256 pixels in size.
	
	The bounding box of the transformed 256x256 image, and the bounding box of the original mask are used to calculate the scale and translation required to transform the dog in the 256x256 image back to its position in the original full-size RGBD image.
	$J_{2Dfull}$ are also transformed using Algorithm \ref{al:imCrop}, producing $J_{2D256}$. 
	Finally, the z-component in $J_{3Dcam}$ is added as the z-component in $J_{2D256}$, giving $J_{3D256}$.
	The x- and y- components of $J_{3D256}$ lie in the range [0,255]. We transform the z-component to lie in the same range by using Algorithm \ref{al:normalisedDepth}. 
	In Algorithm \ref{al:normalisedDepth}, we make two assumptions:
	\begin{enumerate}
		\item The root joint of the skeleton lies within a distance of 8 metres from the camera, the maximum distance detected by a Kinect v2 \citeNew{kinectApiSUPP}
		\item Following Sun et al. \citeNew{sun2018integralSUPP}, the remainder of the joints are defined as offsets from the root joint, normalised to lie within $\pm$ two metres. This is to allow the algorithm to scale to large animals such as horses, etc.
	\end{enumerate}
	
	\begin{algorithm}
		Calculate dog bounding box from binary mask\;
		Apply mask to RGBD image\;
		Crop the image to the bounding box\;
		Make the image square by adding rows/columns in a symmetric fashion\;
		Scale the image to be 256x256 pixels\;
		Add padding to the image bringing the size to 293x293 pixels and rescale the image to 256x256 pixels\;
		\caption{Transform RGBD image for network input}\label{al:imCrop}
		
	\end{algorithm}
	
	\begin{algorithm}
		$rootJoint = J_{3D256}[0]$\;
		\For{$j\in J_{3D256}$}{
			\If{$j == rootJoint$}{
				$rootJointDepth = j[3]$\;
				$j[3] = (min(j[3],8000)/8000)*255$\;
			}
			\Else{
				$j[3] = (j[3] - rootJointDepth) / 2000$\;
				$j[3] = max(min(j[3], 1), -1)$\;
				$j[3] = (j[3] *(255/2)) + (255/2)$\;
			}
		}
		\caption{Normalising joint depth for network input}\label{al:normalisedDepth}
	\end{algorithm}	
	
	\subsection{Pose Prior Model}
\begin{figure}
		\begin{center}
			\includegraphics[width=0.99\linewidth]{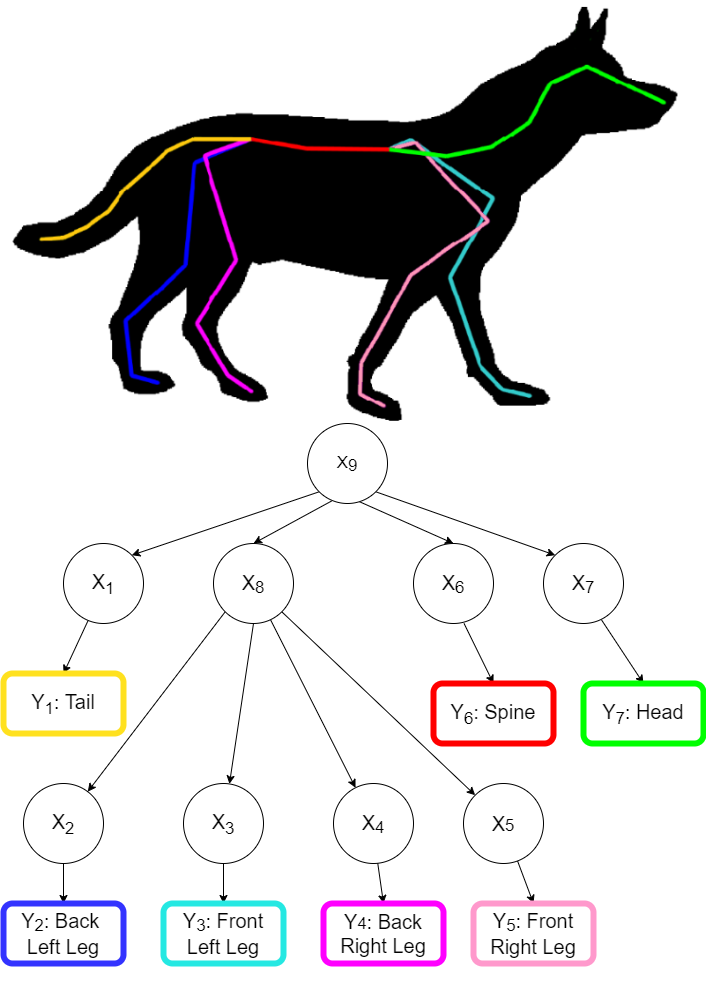}
		\end{center}
		\caption{The structure of our H-GPLVM. Each node $X_i$ produces joint rotations (and translation, if applicable) $Y_i$ for the bones with the corresponding colour.}
		\label{fig:hgpvlm}
	\end{figure}

	\noindent We use a Hierarchical Gaussian Process Latent Variable Model (H-GPLVM) \citeNew{lawrence2007hierarchicalSUPP} to represent high-dimensional skeleton motions lying in a lower-dimensional latent space. 
	Figure \ref{fig:hgpvlm} shows how the structure of the H-GPLVM relates to the structure of the dog skeleton: The latent variable representing the fully body controls the tail, legs, spine, and head variables, while the four legs are further \emph{decomposed} into individual limbs. Equation~\ref{e:jointdistr} shows the corresponding joint distribution.
	
	\begin{align*}
	p(Y_1, Y_2,& Y_3, Y_4, Y_5, Y_6, Y_7) =\\
	&\int P(Y_1 \vert X_1)\ldots   \\
	&\times \int p(Y_2 \vert X_2)\ldots \\
	&\times \int p(Y_3 \vert X_3)\ldots \\
	&\times \int p(Y_4 \vert X_4)\ldots \\
	&\times \int p(Y_5 \vert X_5)\ldots \\
	&\times \int p(Y_6 \vert X_6)\ldots \\
	&\times \int p(Y_7 \vert X_7)\ldots \\
	&\times \int p(X_2, X_3, X_4, X_5 \vert X_8)\ldots \\
	&\times \int p(X_1, X_8, X_6, X_7 \vert X_9) d X_9 \ldots d X_1,\numberthis 
	\label{e:jointdistr}
	\end{align*}
	where $Y_1$ to $Y_7$ are the rotations (and translations, if applicable) of the joints in the tail, back left leg, front left leg, back right leg, front left leg, spine and head respectively and $X_1$ to $X_7$ are the nodes in the model for each respective body part, $X_8$ is the node for all four legs, and $X_9$ is the root node.
	
	Let Y be the motion data matrix of $f$ frames and dimension $d$, $\mathbb{R}^{f\times d}$, containing the data of $Y_1$ to $Y_7$. 
	$K_{x_i}$ is the radial basis function that depends on the $q$-dimensional latent variables $X_i$ that correspond to $Y_i$.
	[$s_i$, $e_{i}$] define the start and end index of columns in $Y$ that contain the data for $Y_i$. $N$ is the normal distribution. Then, 
	\begin{equation}
	p(Y_i \vert X_i)  = \prod_{j=s_i}^{e_i} N ({Y_i}_{[:,j]}|0,K_{x_i}),
	\end{equation}
	where ${Y_i}_{[:,j]}$ denotes the $j$-th column of $Y_i$.

	\subsubsection{\reb{Joint-specific Weights When Fitting the Model}}
	\noindent When fitting the H-GPLVM to the network-predicted joints, each of these joints has an associated weight to guide fitting.
	This is a elementwise-multiplication of two sets of weights, $W_1$ and $W_2$. 
	$W_1$ is user-defined and inspired by the weights used by the Vicon software. 
	Specifically, these are [5,5,5,0.8,0.5,0.8,1,1,1,0.8,0.5,0.8,1,1,1,
	0.8,0.5,0.5,0.8,1,1,0.1,0,0.1,0,0.8,1,1,1,1,0.8,1,1,1,1,1,1,1,1,
	1,1,1,1].
	This has the effect of giving the root and spine the highest weight (5), the end of each limb has a higher weight (1) than the base of the limb (0.8). 
	Each joint in the tail is given equal weights (1).
	As the ears were not included in the model, a weight of 0 was given to the ear tips, and 0.1 given to the base of the ears, in order to slightly influence head rotation.
	
	Prior to the fitting stage, the shape and size of the dog skeleton has either been provided by the user or generated by the PCA shape model.
	The bone lengths $L$ of this skeleton can be calculated.
	For the current frame, we calculate the length of the bones in the skeleton as predicted by the network, $L_{N}$. 
	The deviation from $L$ is then calculated as $abs(L - L_{N})/L$.
	$W_2$ is calculated as the inverse of this deviation, capped to be within the range {[0,1]}.
	
	\section{Evaluation and Results}
	
	\subsection{Ground Truth for BADJA Comparison}
	\noindent In order to compare our results with BADJA \citeNew{biggs2018creaturesSUPP}, we need to calculate the ground truth joints positions of the SMAL skeleton, $S_{SMAL}$.
	Using WrapX \citeNew{wrapxSUPP}, an off-the-shelf mesh registration software package, the neutral mesh of the SMAL model is registered to the neutral mesh of each of the 5 dogs $N_{dog}$, producing the mesh $N_{SMAL}$. 
	We can then represent $N_{SMAL}$ as barycentric coordinates of $N_{dog}$.
	Using these barycentric coordinates, given $N_{dog}$ in a pose, $P_{dog}$, we compute the corresponding $P_{SMAL}$.
	The BADJA joint regressor then produces $S_{SMAL}$ from $P_{SMAL}$.

	The renderer of BADJA \cite{biggs2018creaturesSUPP} mirrors the projection of the predicted skeleton $S_{BADJA}$. 
	This means that for the 2D result, the identity of joints on the left side of $S_{BADJA}$ are swapped with their corresponding paired joints on the right.
	For 3D comparison, we mirror $S_{SMAL}$ with respect to the camera.
	Next, we find the scales, $sc_{SMAL}$ and $sc_{BADJA}$, such that when applied to $S_{SMAL}$ and $S_{BADJA}$ respectively, the head length of both skeletons is 2 units. 
	We apply these scales and also apply $sc_{SMAL}$ to  $S_{GT}$, the ground-truth skeleton that is in our configuration.
	Finally, our predicted skeleton $S_{PRED}$ is scaled to have the same head length as $S_{GT}$.

	\subsection{Comparision to BADJA}
	We include the 2D results when comparing the results of our pipeline with that of Biggs et al. \citeNew{biggs2018creaturesSUPP} in Table \ref{table:badja2d}.
	
	\begin{table}
		\begin{center}
			\resizebox{\linewidth}{!}{%
				\begin{tabular}{|l|l|l|c|c|c|c|}
					\hline
					Dog & Method & Metric & All & Head  	& Body  & Tail\\
					\hline\hline
					\multirow{4}{*}{Dog1} & \multirow{2}{*}{Ours} & MPJPE & \textbf{9.430} &  \textbf{12.788} & \textbf{8.810} & \textbf{8.006} \\
					&	& PCK & \textbf{0.443} & 0.210 & \textbf{0.495} & \textbf{0.514} \\ \cdashline{2-7}
					& \multirow{2}{*}{BADJA\citeNew{biggs2018creaturesSUPP}} & MPJPE & 19.532 & 21.619 & 17.915 & 22.527 \\
					&	& PCK &  0.196 & \textbf{0.214} & 0.225 & 0.089\\
					\hline
					\multirow{4}{*}{Dog2} & \multirow{2}{*}{Ours} & MPJPE & \textbf{11.333} & \textbf{9.098} & 10.703 & \textbf{15.536} \\
					&	& PCK & \textbf{0.424} & \textbf{0.645} & \textbf{0.448} & \textbf{0.128} \\\cdashline{2-7}
					& \multirow{2}{*}{BADJA\citeNew{biggs2018creaturesSUPP}} & MPJPE & 12.154 & 13.163 & \textbf{8.553} & 22.458 \\
					&	& PCK & 0.296 & 0.393 & 0.337 & 0.073 \\
					\hline
					\multirow{4}{*}{Dog3} & \multirow{2}{*}{Ours} & MPJPE & \textbf{9.063} & \textbf{9.152} & \textbf{8.400} & 11.044 \\
					&	& PCK & \textbf{0.450} & \textbf{0.354}  & \textbf{0.492} & \textbf{0.415} \\\cdashline{2-7}
					& \multirow{2}{*}{BADJA\citeNew{biggs2018creaturesSUPP}} & MPJPE & 10.839 & 15.203 &10.597  & \textbf{7.235} \\
					&	& PCK & 0.392 & 0.276 & 0.430 & 0.387 \\
					\hline
					\multirow{4}{*}{Dog4} & \multirow{2}{*}{Ours} & MPJPE & \textbf{11.757} & \textbf{12.968} & \textbf{11.700} & \textbf{10.723} \\
					&	& PCK & \textbf{0.296} & 0.269 & \textbf{0.354} & 0.142 \\ \cdashline{2-7}
					& \multirow{2}{*}{BADJA\citeNew{biggs2018creaturesSUPP}} & MPJPE & 24.936  & 20.964 & 29.222 & 15.439 \\
					&	& PCK & 0.168 & \textbf{0.347} & 0.105 & \textbf{0.189}\\
					\hline 
					\multirow{4}{*}{Dog5} & \multirow{2}{*}{Ours} & MPJPE & \textbf{14.561} & \textbf{14.414} & \textbf{10.523} & 27.329 \\
					&	& PCK & \textbf{0.230} & \textbf{0.189} & \textbf{0.273} & 0.136 \\ \cdashline{2-7}
					& \multirow{2}{*}{BADJA\citeNew{biggs2018creaturesSUPP}} & MPJPE & 20.188 & 15.321 & 21.340 & \textbf{21.436} \\
					&	& PCK & 0.168 & 0.184 & 0.169 & \textbf{0.150}\\
					\hline
			\end{tabular}}
		\end{center}
		\caption[]{2D error results comparing our pipeline and that used in BADJA \citeNew{biggs2018creaturesSUPP} on each of the 5 dogs. Errors are reported relating to the full body or focussed body parts, as shown in Figure \ref{fig:skelComp} of the main paper.}
		\label{table:badja2d}
	\end{table}
	
	\subsection{Applying the Pipeline to Real Kinect Footage}
\begin{figure}
		\begin{center}
			\includegraphics[width=0.99\linewidth]{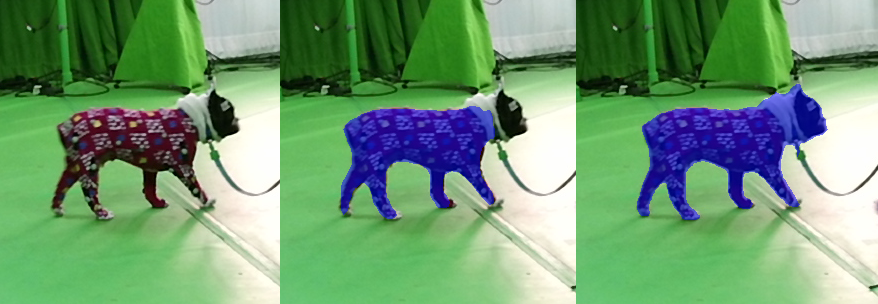}
		\end{center}
		\caption{An example where Deeplab failed to detect a dog in the image (left), the mask as detected by Mask R-CNN (center) and the mask created by Deeplab initialised by the bounding box from Mask R-CNN (right).}
		\label{fig:deeplabVmaskrcnn}
	\end{figure}

	\noindent Running the network on real-world data involves the additional step of generating a mask of the dog from the input image. 
	Two pre-trained networks are used to generate the mask: Mask R-CNN \citeNew{he2017maskSUPP} and Deeplab \citeNew{deeplabv3plus2018SUPP}.
	Both were trained on the COCO dataset \citeNew{lin2014microsoftSUPP} and implemented in Tensorflow.
	During testing, it was found that although Deeplab provided a more accurate mask than Mask R-CNN, it would at times fail to detect any dog in the image, both when the dog is wearing a motion capture suit and when not.
	It would also fail to reliably separate the dog from its handler.
	In our experiments, Mask R-CNN detected the dog in the vast majority of images, although the edge of the mask was not as accurate as that provided by Deeplab.
	Therefore, the image is first processed by Mask R-CNN and the bounding box produced is then used to initialise the input image to Deeplab where it is refined, if possible.
	A comparison of the masks is shown in Figure \ref{fig:deeplabVmaskrcnn}.
	A homography matrix is automatically generated from the Kinect which, when applied to the RGB mask, produces the mask for the depth image.
	
	Table \ref{table:realKinect2d} contains the 2D results when applying our pipeline to real Kinect footage.

	\begin{table}
		\begin{center}
			\resizebox{\linewidth}{!}{%
				\begin{tabular}{|l|l|l|c|c|c|c|}
					\hline
					Dog & Method & Metric & All & Head  	& Body  & Tail\\
					\hline\hline
					\multirow{6}{*}{Dog6} & \multirow{2}{*}{CNN} & MPJPE & 14.754  & \textbf{7.496} & 10.099 & 36.559 \\
					&	& PCK & 0.285 & 0.225 & 0.358 & 0.119 \\ \cdashline{2-7}
					& \multirow{2}{*}{H-GPLVM} & MPJPE & 13.996 & 12.239 & 10.475 & 26.757\\
					&	& PCK &  0.268 	& 0.200 & 0.330 & 0.144\\ \cdashline{2-7}
					& H-GPLVM  & MPJPE & \textbf{6.375} & 7.667 & \textbf{7.764} & \textbf{0.743}\\
					&	(known shape)	& PCK &  \textbf{0.545} & \textbf{0.344} &\textbf{0.528} & \textbf{0.800}\\
					\hline
					\multirow{4}{*}{Dog7} & \multirow{2}{*}{CNN} & MPJPE & \textbf{8.758}  & \textbf{6.461} & \textbf{5.811} & 20.390 \\
					&	& PCK & \textbf{0.456} & \textbf{0.523} & 0.552 & 0.089 \\ \cdashline{2-7}
					& \multirow{2}{*}{H-GPLVM} & MPJPE & 9.533 & 11.383 & 6.391 & \textbf{17.501}\\
					&	& PCK & 0.426 & 0.138 & \textbf{0.576} & \textbf{0.243}\\
					\hline
			\end{tabular}}
		\end{center}
		\caption{2D Error results when using real Kinect images , showing the error result of the network prediction (CNN) and the final pipeline result (H-GPLVM). For Dog6, we also show the error where the shape of the dog mesh and skeleton is known when fitting the H-GPLVM.}
		\label{table:realKinect2d}
	\end{table}

	\subsection{Exploiting Depth Information to Solve Shape}\label{sec:solveShape}
	
	\noindent In this section, different methods for fitting the shape will be described. 
	In all cases, the shape is represented as model parameters to the PCA shape model.
	The results of each method are displayed in Table \ref{table:shapeChange}.
	Each entry in the ``Method" column is described below. 
	
	In general, our pipeline method of solving for shape by referring to bone lengths over a sequence (Original) provided the best results.
	This has the effect of keeping the shape constant for all frames.
	We compare the accuracy of the pipeline when the shape of the dog is allowed to change on a per-frame basis, during the H-GPLVM refinement stage (Method1).
	
	Additionally, we compare the accuracy with our minimisation function takes into account the distance between the mesh produced by the model-generated skeleton to the reconstructed depth points.
	When fitting the model-generated skeleton to the network-predicted joints, we have a one-to-one correspondence as we know the identity of each joint predicted. 
	This is not the case for the vertices on the generated mesh and the reconstructed depth points.
	Matches are made from the generated mesh to the Kinect points, and vice versa using Algorithm \ref{alg:makeMatch}, where the angle threshold is set to 70 degrees, giving the two sets of matches $m_1$ and $m_2$.
	Two tests are performed: the first creates the matches only once during fitting the model (Method2), and the second repeats the matching stage after minimisation up to 3 times provided that the error between the two set of joints reduces by at least 5\% (Method3).
	Finally, two tests were performed with mutual matches only, ie, the matches that appear in both $m_1$ and $m_2$. 
	This match is performed only once (Method 4) or repeated up to 3 times provided that the error between the two set of joints reduces by at least 5\% (Method5).
	
	\begin{algorithm}
		validMatch $= []$\;
		\For{$i=0$ \KwTo $length($sourceMesh$)$}{
			vertexLoc = sourceMesh[$i$]\;
			vertexNormal = sourceNormals[$i$]\;
			nearestMatchInTarget = $knnsearch($vertexLoc, targetMesh$)$\;
			targetLoc = targetMesh[nearestMatchInTarget]\;
			targetNormal = targetNormals[nearestMatchInTarget]\;
			angDiff = $DifferenceInAngles($vertexNormal, targetNormal$)$\;
			\If{angDiff $<$ angleThreshold }{
				validMatch$.append([i$, nearestMatchInTarget$])$\;
			}
		}
		\caption{Creating matches from vertices in the source mesh to the target mesh}
		\label{alg:makeMatch}
	\end{algorithm}

	\begin{table}
		\begin{center}
			\resizebox{\linewidth}{!}{%
				\begin{tabular}{|l|l|l|c|c|c|c|}
					\hline
					Dog & Method & Metric & All & Head  	& Body  & Tail\\
					\hline\hline
					\multirow{12}{*}{Dog6} & \multirow{2}{*}{Original} & MPJPE & 0.667  & \textbf{0.466} & 0.627 & 0.993 \\
					&	& PCK & \textbf{0.873} & \textbf{0.969} & \textbf{0.938} & 0.575 \\ \cdashline{2-7}
					
					& \multirow{2}{*}{Method1} & MPJPE & 0.727 & 0.538 & 0.671 & 1.094\\
					&	& PCK &  0.804 	& 0.887 & 0.848 & 0.581\\ \cdashline{2-7}
					& \multirow{2}{*}{Method2}  & MPJPE & \textbf{0.655} & 0.527 & \textbf{0.599} & \textbf{0.958}\\
					&	& PCK &  0.850 & 0.900 &0.916 & 0.594\\ \cdashline{2-7}
					& \multirow{2}{*}{Method3}  & MPJPE & 0.704 & 0.516 & 0.675 & 0.985\\
					&	& PCK &  0.798 & 0.906 & 0.822 & \textbf{0.613}\\ \cdashline{2-7}
					& \multirow{2}{*}{Method4}  & MPJPE & 0.666 & 0.480  & 0.619 & 1.000 \\
					&	& PCK & 0.843 & 0.938 & 0.892  & 0.594 \\ \cdashline{2-7}
					& \multirow{2}{*}{Method5}  & MPJPE & 0.721 &  0.523 & 0.689  & 1.019 \\
					&	& PCK & 0.787 & 0.912 & 0.816 & 0.569\\
					
					\hline
					\multirow{12}{*}{Dog7} & \multirow{2}{*}{Original} & MPJPE & \textbf{0.557} & \textbf{0.494}  & \textbf{0.471} & \textbf{0.888} \\
					&	& PCK & \textbf{0.922} & \textbf{0.947} & \textbf{0.982} & \textbf{0.711} \\\cdashline{2-7}
					& \multirow{2}{*}{Method1} & MPJPE & 0.902 & 0.740 & 0.784 & 1.436 \\
					&	& PCK & 0.706 & 0.803 & 0.778 & 0.385 \\\cdashline{2-7}
					& \multirow{2}{*}{Method2} & MPJPE & 0.874 & 0.706 & 0.741 & 1.457 \\
					&	& PCK & 0.725 & 0.819 & 0.806 & 0.378 \\\cdashline{2-7}
					& \multirow{2}{*}{Method3} & MPJPE & 0.937 & 0.767 & 0.837 & 1.421 \\
					&	& PCK & 0.655 & 0.763 & 0.704 & 0.395 \\ \cdashline{2-7}
					& \multirow{2}{*}{Method4} & MPJPE & 0.885 & 0.705 & 0.771 & 1.422\\
					&	& PCK & 0.716 & 0.809 & 0.411 & 0.783 \\ \cdashline{2-7}
					& \multirow{2}{*}{Method5} & MPJPE & 0.925 & 0.770 & 0.817 & 1.417\\
					&	& PCK & 0.673 & 0.780 & 0.723 & 0.408 \\
					\hline
			\end{tabular}}
		\end{center}
		\caption[]{3D error results as calculated using PA MPJPE and PA PCK 3D using the original pipeline and the various methods where dog shape can change on a per-frame basis. In general, the best result is achieved when the dog shape is based on bone length of the predicted skeleton and held constant throughout the sequence. A description of each method is provided in Section} \ref{sec:solveShape}.
		\label{table:shapeChange}
	\end{table}
	
	\subsection{Robustness to Occlusions}
	\noindent The training data for the network is free from occlusions.
	To test the pipeline for robustness to occlusions, we apply a mask of a randomly located square.
	This square is 75 pixels in size which is approximately 30\% of the image width/height. 
	As expected, Table \ref{table:testMasked} shows that the results of the pipeline perform worse with the masked image as opposed to the original image.
	However, the H-GPLVM is able to reduce the error of the joint locations.
	
	\begin{table}
		\begin{center}
			\resizebox{\linewidth}{!}{%
				\begin{tabular}{|l|l|l|c|c|c|c|}
					\hline
					Dog & Method & Metric & All & Head  	& Body  & Tail\\
					\hline\hline
					\multirow{6}{*}{Dog6} & \multirow{2}{*}{CNN} & MPJPE & 1.100  & 0.811 & 1.085 & 1.436 \\
					&	& PCK & 0.584 & \textbf{0.819} & 0.554 & 0.444 \\ \cdashline{2-7}
					& \multirow{2}{*}{H-GPLVM} & MPJPE & \textbf{1.005} & \textbf{0.746} & \textbf{1.027} & \textbf{1.193}\\
					&	& PCK &  \textbf{0.606} & 0.794 & \textbf{0.596} & \textbf{0.450}\\ \cdashline{2-7}
					& \multirow{2}{*}{Original} & MPJPE & 0.667 & 0.466 & 0.627 & 0.993\\
					&	& PCK &  0.873 & 0.969 & 0.938 & 0.575\\
					\hline
					
					\multirow{6}{*}{Dog7} & \multirow{2}{*}{CNN} & MPJPE & 0.814  & \textbf{0.609} & 0.760 & 1.189 \\
					&	& PCK & \textbf{0.769} & \textbf{0.868} & 0.791 & 0.602 \\ \cdashline{2-7}
					& \multirow{2}{*}{H-GPLVM} & MPJPE & \textbf{0.781} & 0.673 & \textbf{0.711} & \textbf{1.110}\\
					&	& PCK & 0.768 & 0.816 & \textbf{0.801} & \textbf{0.618}\\ \cdashline{2-7}
					& \multirow{2}{*}{Original} & MPJPE & 0.557 & 0.494 & 0.471 & 0.888\\
					&	& PCK &  0.922 & 0.947 & 0.982 & 0.711\\
					\hline
			\end{tabular}}
		\end{center}
		\caption{3D Error results of PA MPJPE and PA PCK 3D when using real Kinect images that have been randomly masked, where each skeleton is scaled such that the head has length of two units. We give the errors of the two stages of the pipeline, showing that the H-GPVLM can improve the network result. The original errors of the pipeline are shown for ease of comparison and are not highlighted if more accurate results were achieved.}
		\label{table:testMasked}
	\end{table}
	
	\subsection{Comparison With Other Networks}
	
	\noindent We compare the network result of our pipeline, which uses the stacked-hourglass network of Newell et al. \citeNew{newell2016stackedSUPP}, with the networks of Sun et al. \citeNew{sun2018integralSUPP} and Moon et al. \citeNew{moon2018v2vSUPP}.
	The networks were given the given the same training, validation and test data and trained for the same number of epochs. 
	Tables \ref{table:networkComp_2d} and \ref{table:networkComp_3d} show that the network of Newell et al. \citeNew{newell2016stackedSUPP} produced more accurate predictions in both 2D and 3D.
	
	\begin{table}
		\begin{center}
			\resizebox{\linewidth}{!}{%
				\begin{tabular}{|l|l|l|c|c|c|c|}
					\hline
					Dog & Network & Metric & All & Head  	& Body  & Tail\\
					\hline\hline
					\multirow{6}{*}{Dog6} & \multirow{2}{*}{Newell et al.} & MPJPE & \textbf{14.754}  & \textbf{7.496} & \textbf{10.099} & 36.559 \\
					&	& PCK & \textbf{0.285} & \textbf{0.225} & \textbf{0.358} & 0.119 \\ \cdashline{2-7}
					& \multirow{2}{*}{Sun et al.} & MPJPE & 30.219 & 37.329 & 27.602 & 29.513\\
					&	& PCK &  0.078 	& 0.050 & 0.076 & 0.119\\ \cdashline{2-7}
					& \multirow{2}{*}{Moon et al.} & MPJPE & 16.791 & 14.148 & 14.779  &  \textbf{26.383} \\
					&	& PCK &  0.155 	& 0.160  & 0.031  & \textbf{0.192} \\
					\hline
					\multirow{6}{*}{Dog7} & \multirow{2}{*}{Newell et al.} & MPJPE & \textbf{8.758}  & \textbf{6.461} & \textbf{5.811} & 20.390 \\
					&	& PCK & \textbf{0.456} & \textbf{0.523} & \textbf{0.552} & 0.089 \\ \cdashline{2-7}
					& \multirow{2}{*}{Sun et al.} & MPJPE & 11.904 & 10.381 & 7.870 & 26.412\\
					&	& PCK &  0.364 	& 0.345 & 0.411 & \textbf{0.243}\\ \cdashline{2-7}
					& \multirow{2}{*}{Moon et al.} & MPJPE & 14.693 & 10.593 & 15.479  & \textbf{17.358} \\
					&	& PCK &   0.239	& 0.321  &  0.245 & 0.115 \\
					\hline
			\end{tabular}}
		\end{center}
		\caption[]{2D MPJPE and PCK error results when using real Kinect images as produced by the networks of Newell et al. \citeNew{newell2016stackedSUPP}, Sun et al. \citeNew{sun2018integralSUPP} and Moon et al. \citeNew{moon2018v2vSUPP}. In general, Newell et al. \citeNew{newell2016stackedSUPP} performs best.}
		\label{table:networkComp_2d}
	\end{table}
	
	\begin{table}
		\begin{center}
			\resizebox{\linewidth}{!}{%
				\begin{tabular}{|l|l|l|c|c|c|c|}
					\hline
					Dog & Network & Metric & All & Head  	& Body  & Tail\\
					\hline\hline
					\multirow{6}{*}{Dog6} & \multirow{2}{*}{Newell et al.} & MPJPE & \textbf{0.866}  & \textbf{0.491} & \textbf{0.776} & 1.523 \\
					&	& PCK& \textbf{0.745} & \textbf{0.956} & \textbf{0.780} & 0.425 \\ \cdashline{2-7}
					& \multirow{2}{*}{Sun et al.} & MPJPE & 1.594 & 1.561 & 1.723 & 1.341\\
					&	& PCK &  0.279 	& 0.300 & 0.340 & 0.250\\ \cdashline{2-7}
					& \multirow{2}{*}{Moon et al.} & MPJPE & 0.896 & 0.879 &   0.912 &  \textbf{0.867}\\
					&	& PCK &  0.715 	& 0.685  & 0.714  & \textbf{0.756} \\
					\hline
					
					\multirow{6}{*}{Dog7} & \multirow{2}{*}{Newell et al.} & MPJPE & \textbf{0.563}  & \textbf{0.364} & \textbf{0.507} & 0.939 \\
					&	& PCK & \textbf{0.907} & \textbf{0.993} & \textbf{0.943} & 0.707 \\ \cdashline{2-7}
					& \multirow{2}{*}{Sun et al.} & MPJPE & 0.889 & 0.698 & 0.810 & 1.372\\
					&	& PCK &  0.734 	& 0.821 & 0.743 & 0.595\\ \cdashline{2-7}
					& \multirow{2}{*}{Moon et al.} & MPJPE & 0.901 & 0.667 & 1.017  & \textbf{0.832} \\
					&	& PCK &  0.715	& 0.834  &  0.649 & \textbf{0.770} \\
					
					\hline
			\end{tabular}}
		\end{center}
		\caption[]{3D Error results of PA MPJPE and PA PCK 3D when using real Kinect images, where the ground-truth skeleton is scaled such that the head has length of two units. We show the errors for the networks of Newell et al. \citeNew{newell2016stackedSUPP}, Sun et al. \citeNew{sun2018integralSUPP} and Moon et al. \citeNew{moon2018v2vSUPP}, with Newell et al performing best.}
		\label{table:networkComp_3d}
	\end{table}
	
	The method of Moon et al. \citeNew{moon2018v2vSUPP} predicts 3D joint positions based on the voxel representation of the depth image.
	The author's pipeline first uses the DeepPrior++ network of Oberweger and Lepetit \citeNew{oberweger2017deepprior++SUPP} to predict the location of a reference point based on the centre of mass of the voxels.
	This reference point used to define the other joints in the skeleton and is more feasible to predict than the root of the skeleton itself.
	Due to memory and time constraints, the training data for this network contained the synthetic jump sequence of a single dog as seen by 28 cameras.
	
	To test the result of this network, we calculate the mean euclidean distance from the reference point to the root of the ground-truth skeleton across all frames . 
	We compare this to the distance from the center of mass of the voxels to the root.
	First we test the network on a single camera of a synthetic trot sequence of the training dog.
	The mean distance for the reference point was 302.64mm and mean distance for the center of mass was 302.55mm.
	Next we tested the network on two real Kinect sequences where again the reference point increased the error of the center of mass point by approximately 0.1mm.
	As a result, the center of mass was used as the reference point for each image when training the network of Moon et al. \citeNew{moon2018v2vSUPP}, rather than that predicted by DeepPrior++.
	
	\subsection{Comparison of Our Shape Model with SMAL}

	\noindent As the skeleton configuration of the two shape models are different, the SMAL model cannot be directly fit to network-predicted joints.
	Instead, to compare the models, we fit each model to the neutral dog mesh and skeleton of each dog in the set of Dog1-Dog5. 
	For each dog, the average SMAL mesh is registered to the original dog mesh and the corresponding joint locations are calculated using the SMAL joint regressor. 
	A different version of our shape model is created for each test where the information for the test dog is removed from the shape model.
	
	We aim to find the shape parameters for each model that produces the mesh that most accurately represents each dog.
	As the scale of the SMAL model differs to the dog meshes, the overall scale of both models is also optimised in this process along with the shape parameters.
	For each model, we report the error result as the mean euclidean distance from each joint in the skeleton as produced by the model and the ground-truth joint in millimetres.
	We report the same error for each vertex in the meshes.
	These are shown in each row of Table \ref{table:solveShape}.
	We perform tests where the models fit to only joint information (the first row of Table \ref{table:solveShape}), fit to only vertex information (the second row) and both joint and vertex information (the third row).
	
	This assumes that the pose of the model and that of the test dog are identical, which may not the case.
	As such, we then performed tests where the pose can change, i.e. we now solve for scale, shape parameters and pose parameters when fitting the model.
	The steps described above are repeated, and the results are reported in the final three rows of Table \ref{table:solveShape}.
	
	In general, the SMAL model achieved better results when the pose of the dog was fixed, whereas our model achieved better results when the pose was allowed to move.
	We believe this is due to each animal in the SMAL model having a similar neutral pose to each other whereas the neutral pose in our model is dog-specific.
	
	
	
	\begin{table} 
		\begin{center}
			\resizebox{\linewidth}{!}{%
				\begin{tabular}{|l|l|c|c|c|c|}
					\hline
					\multirow{2}{*}{ } & \multirow{2}{*}{Model Fit To} &  \multicolumn{2}{|c|}{Errors - Ours} &  \multicolumn{2}{|c|}{Errors - SMAL}\\	
					& & joints & mesh & joints & mesh \\
					
					\hline\hline
					\multirow{3}{*}{Fixed pose} &joints & 45.458 & 26.819 & \textbf{37.824} & \textbf{23.182} \\
					&mesh & 44.050 & \textbf{69.923} & \textbf{23.221} & 72.220 \\
					&joints \& mesh & 45.190 & 26.915 & \textbf{37.636} & \textbf{23.242} \\	
					\hline
					
					\multirow{3}{*}{Solved pose} &joints & \textbf{18.331} & \textbf{10.430} & 23.582 & 17.925 \\
					&mesh & \textbf{7.225} & \textbf{25.649} & 11.138 & 56.058 \\
					&joints \& mesh & \textbf{17.255} & \textbf{10.1585} & 22.175 & 14.689 \\	
					\hline
					
				\end{tabular}
			}
		\end{center}
		\caption{Given the corresponding configuration of ground-truth mesh and joint locations, for each dog in the set Dog1-Dog5, we find the global scale and model parameters of our shape model and the SMAL model that best fits to just the joint locations, just the mesh, or both the joints and mesh (rows 1-3). This test is repeated when finding the global scale, model parameters and skeleton pose parameters (rows 4-6). Errors are reported as the mean euclidean distance in millimetres for either each joint in the skeleton or each vertex in the mesh. SMAL achieves better results for a fixed pose, and our model achieved better results when the pose of the skeleton was allowed to change.
		}
		\label{table:solveShape}
	\end{table}

	{\small
		
		\bibliographystyleNew{ieee_fullname}
		\bibliographyNew{egbib}
		
	}

\end{document}